\documentclass{article}

\usepackage{microtype}
\usepackage{graphicx}
\usepackage{subfigure}
\usepackage{booktabs} % for professional tables

\usepackage{hyperref}

\usepackage[accepted]{sty}
\usepackage{graphicx}
\usepackage{pythonhighlight}

\usepackage{times}
\usepackage{helvet}
\usepackage{courier}
\usepackage{subfigure} 
\usepackage{cancel}
\RequirePackage{amsthm,amsmath}
\usepackage{graphicx, amssymb, mathrsfs,amsfonts,url, bm}

\newcommand{\rms}{\text{RMS}}

\newcommand{\real}{\mathbb{R}}

\newcommand{\ba}{\bm{a}}

\newcommand{\bb}{\bm{b}}

\newcommand{\bc}{\bm{c}}

\newcommand{\be}{\bm{e}}

\newcommand{\bg}{\bm{g}}

\newcommand{\bH}{\bm{H}}

\newcommand{\bn}{\bm{n}}

\newcommand{\bs}{\bm{s}}

\newcommand{\bx}{\bm{x}}

\newcommand{\titlethis}{AdaSmooth: An Adaptive Learning Rate Method based on Effective Ratio}

\icmltitlerunning{\titlethis}

\begin{document}

\twocolumn[
\icmltitle{\titlethis}

\begin{icmlauthorlist}
\icmlauthor{Jun Lu}{te}
\end{icmlauthorlist}

\icmlaffiliation{te}{Correspondence to: Jun Lu $<$jun.lu.locky@gmail.com$>$. Copyright 2022 by the author(s)/owner(s). April 2nd, 2022}

%\icmlcorrespondingauthor{Jun Lu}{jun.lu.locky@gmail.com}

\vskip 0.3in
]



\printAffiliationsAndNotice{}  % leave blank if no need to mention equal contribution

\begin{abstract}

It is well known that we need to choose the hyper-parameters in Momentum, AdaGrad, AdaDelta, and other alternative stochastic optimizers. While in many cases, the hyper-parameters are tuned tediously based on experience becoming more of an art than science. We present a novel per-dimension learning rate method for gradient descent called AdaSmooth. The method is insensitive to hyper-parameters thus it requires no manual tuning of the hyper-parameters like Momentum, AdaGrad, and AdaDelta methods. We show promising results compared to other methods on different convolutional neural networks, multi-layer perceptron, and alternative machine learning tasks. Empirical results demonstrate that AdaSmooth works well in practice and compares favorably to other stochastic optimization methods in neural networks. 

\end{abstract}
\section{Introduction}
Over the years, stochastic gradient-based optimization has become a core method in many fields of science and engineering such as computer vision and automatic speech recognition processing \citep{krizhevsky2012imagenet, hinton2012deep, graves2013speech}. Stochastic gradient descent (SGD) and deep neural network (DNN) play a core role in training stochastic objective functions. When a new deep neural network is developed for
a given task, some hyper-parameters related to the training of the network must be chosen heuristically. For each possible combination of structural hyper-parameters, a new network is typically
trained from scratch and evaluated over and over again. While much progress has been made on hardware (e.g. Graphical Processing Units) and software (e.g. cuDNN) to speed up the training time of a single structure
of a DNN, the exploration of a large set of possible structures remains very slow making the need of a stochastic optimizer that is insensitive to hyper-parameters.

\subsection{Gradient Descent}
Gradient descent (GD) is one of the most popular algorithms to perform optimization and by far the
most common way to optimize machine learning tasks. And this is particularly true for optimizing neural networks. 
The neural networks or machine learning in general find the set of parameters $\bx\in \real^d$ %(a.k.a., weights) 
in order to optimize an objective function $L(\bx)$. The gradient descent finds a sequence of parameters 
\begin{equation}
\bx_1, \bx_2, \ldots, \bx_T,
\end{equation}
such that when $T\rightarrow \infty$, the objective function $L(\bx_T)$ achieves the optimal minimum value.
At each iteration $t$, a step $\Delta \bx_t$ is applied to change the parameters. Denoting the parameters at the $t$-th iteration as $\bx_t$. 
Then the update rule becomes 
\begin{equation}
\bx_{t+1} = \bx_t + \Delta \bx_t.
\end{equation}

The most naive method of stochastic gradient descent is the vanilla update: the parameter moves in the opposite direction of the gradient which finds the steepest descent direction since the gradients are orthogonal to level curves (a.k.a., level surface, see Lemma 16.4 in \citet{lu2022matrix}): 
\begin{equation}
\Delta \bx_{t} = -\eta \bg_t= -\eta \frac{\partial L(\bx_t)}{\partial \bx_t} = -\eta \nabla L(\bx_t),
\end{equation}
where the positive value $\eta$ is the learning rate and depends on specific problems, and $\bg_t=\frac{\partial L(\bx^t)}{\partial \bx^t} \in 
\real^d$ is the gradient of the parameters.
% i.e., the update rule becomes 
%\begin{equation}
%\bx_{t+1} = \bx_t -\eta \bg_t,
%\end{equation}
%i.e., moving in the direction of negative of the gradient. 
The learning rate $\eta$ controls how large of a step to take in the direction of negative gradient so that we can reach a (local) minimum.
While if we follow the negative gradient of a single sample or a batch of samples iteratively, the local estimate of the direction can be obtained and is known as the stochastic gradient descent (SGD) \citep{robbins1951stochastic}.
In the SGD framework, the objective function is stochastic that is composed of a sum of subfunctions evaluated at different subsamples of the data.

%The drawback of vanilla update is that it is easy to get stuck in local minimum \citep{rutishauser1959theory}.

For a small step-size, gradient descent makes a monotonic improvement at every iteration. Thus, it always converges, albeit to a local minimum. However, the speed of the vanilla GD method is usually slow, while it can take an exponential rate when the curvature condition is poor. While choosing higher than this rate may cause the procedure to diverge in terms of the objective function. Determining a good learning rate (either global or per-dimension) becomes more of an art than science for many problems. Previous work has been done to alleviate the need for selecting a global learning rate \citep{zeiler2012adadelta}, while it is still sensitive to other hyper-parameters.

%Drawback (from AdaDelta): Choosing higher than this rate can cause the system
%to diverge in terms of the objective function, and choosing
%this rate too low results in slow learning. Determining a good
%learning rate becomes more of an art than science for many
%problems.

The main contribution of this paper is to propose a novel stochastic optimization method which is insensitive to different choices of hyper-parmeters resulting in adaptive learning rates, and naturally performs a form of step size annealing. We propose the AdaSmooth algorithm to both increase optimization efficiency and out-of-sample accuracy. While previous works propose somewhat algorithms that are insensitive to the global learning rate (e.g., \citep{zeiler2012adadelta}), the methods are still sensitive to hyper-parameters that influence per-dimension learning rates. The proposed AdaSmooth, a method for efficient stochastic optimization that only requires first-order gradients and (accumulated) past update steps with little memory requirement, allows flexible and adaptive per-dimension learning rates. Meanwhile, the method is memory efficient and easy to implement. Our method is designed to combine the advantages of the following methods: AdaGrad \citep{duchi2011adaptive}, which works well with sparse gradients, RMSProp \citep{hinton2012neural}, which works well in online and non-stationary settings, and AdaDelta \citep{zeiler2012adadelta}, which is less sensitive in global learning rate.
%while important connections to these stochastic optimization are discussed in Sec. 

%\subsection{Learning Rate Annealing}\label{section:learning-rate-annealing}
%The aforementioned gradient descent needs to choose a proper learning rate. 
%However, choosing a proper learning rate can be difficult. A learning rate that is too small leads to
%painfully slow convergence, while a learning rate that is too large can hinder convergence
%and cause the loss function to fluctuate around the minimum or even to diverge.
%
%
%(AdaDelta) When gradient descent nears a minima in the cost surface, the parameter values can oscillate back and forth around
%the minima. One method to prevent this is to slow down the
%parameter updates by decreasing the learning rate. This can
%be done manually when the validation accuracy appears to
%plateau. Alternatively, learning rate schedules have been proposed to automatically anneal the learning rate based on
%how many epochs through the data have been done \citep{robbins1951stochastic}. These approaches typically add additional hyper-parameters to control
%how quickly the learning rate decays.

\section{Related Work}

There are several variants of gradient descent to use heuristics for estimating a good learning rate at each iteration of the progress. These methods either attempt to accelerate learning when suitable or to slow down learning near a local minima \citep{zeiler2012adadelta, kingma2014adam, ruder2016overview}.
 
%which differ in how much data we use to compute the
%gradient of the objective function. Depending on the amount of data, we make a trade-off between
%the accuracy of the parameter update and the time it takes to perform an update.

\begin{figure}[h]
\centering  
\vspace{-0.35cm} 
\subfigtopskip=2pt 
\subfigbottomskip=2pt 
\subfigcapskip=-5pt 
\subfigure[Optimization without Momentum. A higher learning rate
may result in larger parameter updates in dimension across the valley (direction of $x_2$) which could lead
to oscillations back as forth across the valley.]{\label{fig:momentum_gd}
	\includegraphics[width=0.99\linewidth]{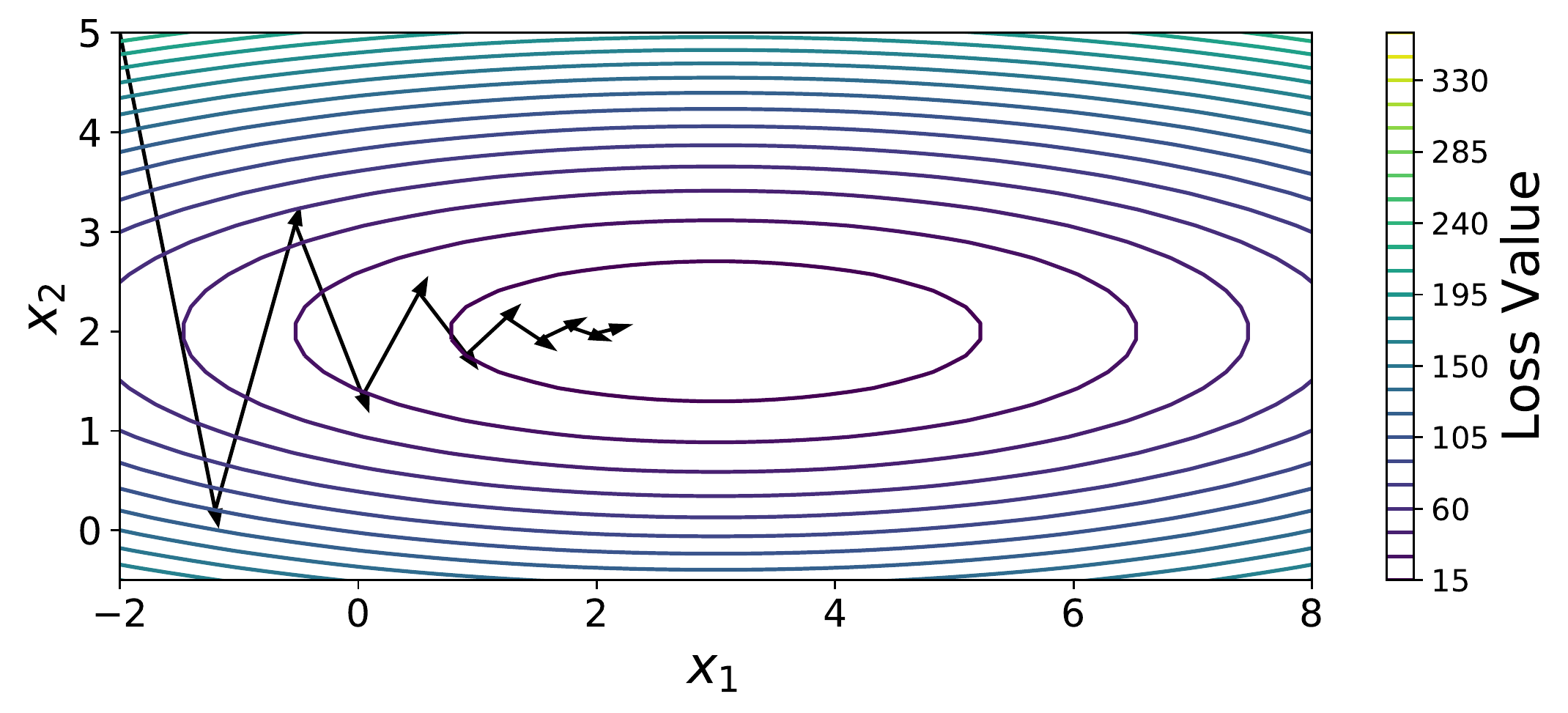}}
\subfigure[Optimization with Momentum. Though the gradients along the valley (direction of $x_1$) are much smaller than the gradients across the valley (direction of $x_2$), they are
typically in the same direction and thus the momentum term
accumulates to speed up movement, dampen oscillations and cause us to barrel through narrow valleys, small humps and (local) minima.]{\label{fig:momentum_mum}
	\includegraphics[width=0.99\linewidth]{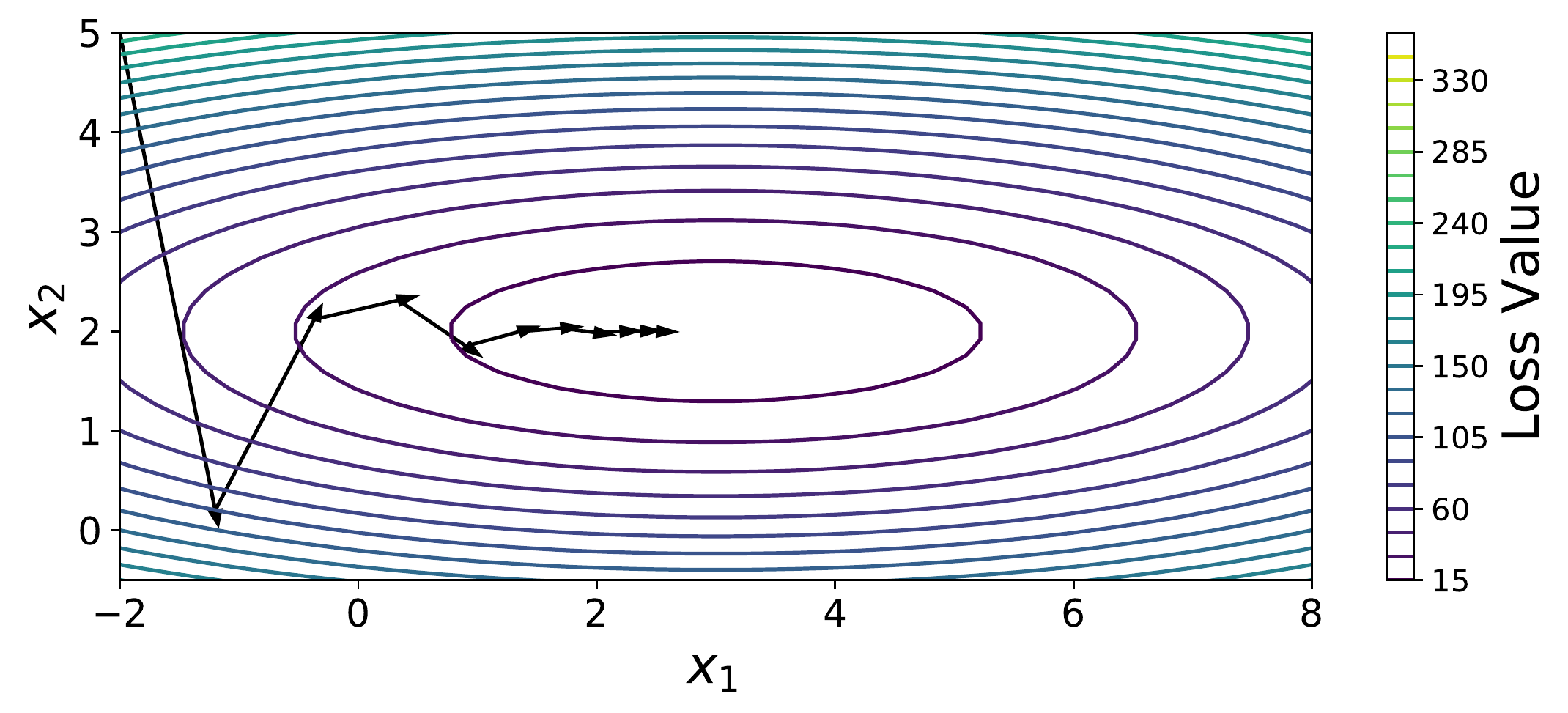}}
\caption{A 2-dimensional convex function $L(\bx)=2(x_1-3)^2 + 20(x_2-2)^2+5$ and $\frac{\partial L(\bx)}{\partial \bx}=(4x_1-12, 8x_2-16)^\top$. Starting point to descent is $(-2, 5)^\top$.}
\label{fig:momentum_gd_compare}
\end{figure}

\subsection{Momentum }
If the cost surface is not spherical, learning can be quite slow because the learning rate must be kept small to prevent divergence along the steep curvature directions \citep{rumelhart1986learning, qian1999momentum, sutskever2013importance}. 
The SGD with Momentum (that can be applied to full batch or mini-batch learning) attempts to use previous step to speed up learning when suitable such that it enjoys better converge rates on deep networks.
The main idea behind the Momentum method is to speed up the learning along dimensions where the gradient consistently point in the same direction; and to slow the pace along dimensions in which the sign of the gradient continues to change. Figure~\ref{fig:momentum_gd} shows a set of updates for vanilla GD where we can find the update along dimension $x_1$ is consistent; and the move along dimension $x_2$ continues to change in a zigzag pattern. The GD with Momentum keeps track of past parameter updates with an exponential decay, and the update method has the following step:
\begin{equation}
\begin{aligned}
	\Delta \bx_t &= \rho \Delta \bx_{t-1} - \eta \frac{\partial L(\bx_t)}{\partial \bx_t},\\
\end{aligned}
\end{equation}
where the algorithm remembers the latest update and adds it to present update by multiplying a parameter $\rho$ called \textit{momentum parameter}. 
%That is, the modification of the weight vector at the current time step $t$ depends on both the current gradient (negative) and the weight change of the previous step $t-1$. 
That is, the amount we change the parameter is proportional to the negative gradient plus the previous weight change; the added momentum term acts as both a smoother and an accelerator.
The momentum parameter $\rho$ works as a \textit{decay constant} where $\Delta\bx_1$ may has effect on $\Delta \bx_{100}$; however, its effect is decayed by this decay constant. In practice, the momentum parameter $\rho$ is usually set to be 0.9 by default.
Momentum simulates the concept inertia in physics. It means that in each iteration, the update mechanism is not 
only related to the gradient descent, which refers to the dynamic term, but also maintains a component 
which is related to the direction of last update iteration, which refers to the momentum.

The momentum works extremely better in ravine-shaped loss curve. Ravine is an area, where the surface curves are much more steeply in one dimension than in another (see the loss curve in Figure~\ref{fig:momentum_gd_compare}, i.e., a long narrow valley). 
Ravines are common near local minima in deep neural networks and vanilla SGD has trouble navigating them. As shown by the toy example in Figure~\ref{fig:momentum_gd}, SGD will tend to oscillate across the narrow ravine since the negative gradient will point down one of the steep sides rather than along the ravine towards the optimum. Momentum helps accelerate gradients in the correct direction (Figure~\ref{fig:momentum_mum}).

\subsection{AdaGrad}
The learning rate annealing procedure modifies a single global learning rate that applies to all dimensions of the parameters \citep{smith2017cyclical}. \citet{duchi2011adaptive} proposed a method called AdaGrad where the learning rate is updated on a per-dimension basis. 
The learning rate for each parameter depends on the history of gradient updates of that parameter in a way such that parameters with a scarce history of updates are updated faster using a larger learning rate. In other words, parameters that have not been updated much in the past are more likely to have higher learning rates now. Denoting the element-wise vector multiplication between $\ba$ and $\bb$ by $\ba\odot\bb$, formally, the AdaGrad has the following update step:
\begin{equation}
\begin{aligned}
\Delta \bx_t &= -  \frac{ \eta}{ \sqrt{\sum_{\tau=1}^t \bg_{\tau}^2 +\epsilon} }\odot \bg_{t} ,
\end{aligned}
\end{equation}
where $\epsilon$ is a smoothing term to better condition the division, $\eta$ is a global learning rate shared by all dimensions, $\bg_\tau^2$ indicates the element-wise square $\bg_\tau\odot \bg_\tau$, and the denominator computes the $l$2 norm of sum of all previous squared gradients in a per-dimension fashion. Though the $\eta$ is shared by all dimensions, each dimension has its own dynamic learning rate controlled by the $l$2 norm of accumulated gradient magnitudes. Since this dynamic learning rate grows with the inverse of the accumulated gradient magnitudes, larger gradient magnitudes have smaller learning rates and smaller absolute values of gradients have larger learning rates. Therefore, the accumulated gradient in the denominator has the same effects as the learning rate annealing.

One pro of the AdaGrad method is that it partly eliminates the need to tune the learning rate %which is 
controlled by the accumulated gradient magnitude. 
However, AdaGrad's main weakness is its accumulation of the squared gradients in the denominator. Since every added term is positive, the accumulated sum 
%the continual accumulation of squared gradients
keeps growing or exploding during every training step. This in turn causes the per-dimension learning rate to shrink and eventually decrease throughout training and become infinitesimally small, eventually falling to zero and stopping training any more. 
Moreover, since the magnitudes of gradients are factored out in
AdaGrad, this method can be sensitive to the initialization
of the parameters and the corresponding gradients. If the initial magnitudes of the gradients are large or infinitesimally huge, the per-dimension learning rates will be low for the remainder of training. 
This can be partly combated by increasing the global learning rate, making the AdaGrad method sensitive to the choice of learning rate. 
Further, since AdaGrad assumes the parameter with fewer updates should favor a larger learning rate; and one with more movement should employ a smaller learning rate. This makes it
consider only  the information from squared gradients, or the absolute value of the gradients. And thus AdaGrad does not include information from total move (i.e., the sum of updates; not the sum of absolute updates).

To be more succinct, AdaGrad has the following main drawbacks:
1) the continual decay of learning rates throughout training;
2) the need for a manually selected global learning rate;
3) considering only the absolute value of gradients.

\subsection{AdaDelta}

AdaDelta is an extension of AdaGrad that overcomes the main weakness of AdaGrad \citep{zeiler2012adadelta}. The original idea of AdaDelta is simple: it restricts the window of accumulated past gradients to some fixed size $w$ rather than $t$ (i.e., current time step). 
However, since storing $w$ previous squared gradients is inefficient, the AdaDelta introduced in \citet{zeiler2012adadelta} implements this accumulation as an exponentially decaying average of the squared gradients. This is very similar to the idea of momentum term (or decay constant).

\subsubsection{AdaDelta: Form 1 (RMSProp)}\label{section:adawin}
We first discuss the exact form of the window AdaGrad (we here call it AdaGradWin for short).
Assume at time $t$ this running average is $E[\bg^2]_t$ then we compute:
\begin{equation}\label{equation:adagradwin}
E[\bg^2]_t = \rho E[\bg^2]_{t-1} + (1 - \rho) \bg_{t}^2,
\end{equation}
where $\rho$ is a decay constant similar to that used in the momentum method and $\bg_t^2$ indicates the element-wise square $\bg_t\odot \bg_t$.

As Eq~\eqref{equation:adagradwin} is just the root mean squared (RMS) error criterion of the gradients, we can replace
it with the criterion short-hand.
Let $\rms[\bg]_t = \sqrt{E[\bg^2]_t + \epsilon}$, where again a constant $\epsilon$ is added to better condition the denominator. Then the resulting step size can be obtained as follows:
\begin{equation}\label{equation:rmsprop_update}
\Delta \bx_t=- \frac{\eta}{\rms[\bg]_t}  \odot \bg_{t},
\end{equation}
where again $\odot$ is the element-wise vector multiplication.
%Similar to momentum, $\rho$ in AdaGradWin is also set to 0.9 by default. 

As aforementioned, the form in Eq~\eqref{equation:adagradwin} is originally from the exponential moving average (EMA). In the original form of EMA, $1-\rho$ is also known as the smoothing constant (SC) where the SC can be written as $\frac{2}{N+1}$ and the period $N$ can be thought of as the number of past values to do the moving average calculation \citep{lu2022exploring}:
\begin{equation}\label{equation:ema_smooting_constant}
\text{SC}=1-\rho = \frac{2}{N+1}.
\end{equation}
The above Eq~\eqref{equation:ema_smooting_constant} links different variables: the decay constant $\rho$, the smoothing constant (SC), and the period $N$.
If $\rho=0.9$, then $N=19$. That is, roughly speaking, $E[\bg^2]_t $ at iteration $t$ is approximately equal to the moving average of past 19 squared gradients and the current one (i.e., moving average of 20 squared gradients totally).
The relationship in Eq~\eqref{equation:ema_smooting_constant} though is not discussed in \citet{zeiler2012adadelta}, it is important to decide the lower bound of the decay constant $\rho$. Typically, a time period of $N=3$ or 7 is thought to be a relatively small frame making the lower bound of decay constant $\rho=0.5$ or 0.75; when $N\rightarrow \infty$, the decay constant $\rho$ approaches $1$.

%Roughly speaking, $E[g_i^2]_t$ at iteration $t$ is a moving average of past $39$ squared gradients and the current one (i.e., moving average of 40 squared gradients totally). 
However, we can find that the AdaGradWin still only considers the absolute value of gradients and a fixed number of past squared gradients is not flexible which can cause a small learning rate near (local) minima as we will discuss in the sequel. 

\paragraph{RMSProp} The AdaGradWin is actually the same as the RMSProp method developed independently by Geoff Hinton in \citet{hinton2012neural} both of which are stemming from the need to resolve AdaGrad's radically diminishing per-dimension learning rates. \citet{hinton2012neural} suggests $\rho$ to be set to 0.9 and the global learning rate $\eta$ to be $0.001$ by default.

\subsubsection{AdaDelta: Form 2}
\citet{zeiler2012adadelta} shows the units of the step size shown above do not match (so as the vanilla SGD, the momentum, and the AdaGrad). To overcome this weakness, from the correctness of the second order method, the author considers to rearrange Hessian to determine the quantities involved.
It is well known that, though the calculation of Hessian or approximation to the Hessian matrix is a tedious and computationally expensive task, its curvature information is useful for optimization, and the units in Newton's method are well matched. Given the Hessian matrix $\bH$, the update step in Newton's method can be described as follows \citep{becker1988improving, dauphin2014identifying}:
\begin{equation}
\Delta\bx_t \propto - \bH^{-1} \bg_t \propto \frac{\frac{\partial L(\bx_t)}{\partial \bx_t}}{\frac{\partial^2 L(\bx_t)}{\partial \bx^2}}.
\end{equation}
This implies 
\begin{equation}
\frac{1}{\frac{\partial^2 L(\bx_t)}{\partial \bx_t^2}} = \frac{\Delta \bx_t}{\frac{\partial L(\bx_t)}{\partial \bx_t}},
\end{equation}
i.e., the units of the Hessian matrix can be approximated by the right-hand side term of the above equation. Since the RMSProp update in Eq~\eqref{equation:rmsprop_update} already involves $\rms[\bg]_t$ in the denominator, i.e., the units of the gradients. Putting another unit of the order of $\Delta \bx_t$ in the numerator can match the same order as Newton's method. To do this, define another exponentially decaying average of the update steps:
\begin{equation}
\begin{aligned}
\rms[\Delta \bx]_t &= 	\sqrt{E[\Delta \bx^2]_t } \\
&= \sqrt{\rho E[\Delta \bx^2]_{t-1} + (1 - \rho) \Delta \bx_{t}^2  }.
\end{aligned}
\end{equation}
Since $\Delta \bx_t$ for the current iteration is not known and the curvature can be assumed to be locally smoothed making it suitable to approximate $\rms[\Delta \bx]_t$ by $\rms[\Delta \bx]_{t-1}$. 
%\[\Delta \bx_t = \frac{\frac{\partial L(\bx_t)}{\partial \bx_t}}{\frac{\partial^2 L(\bx_t)}{\partial \bx^2}} \rightarrow \frac{1}{\frac{\partial^2 L(\bx_t)}{\partial \bx^2}} = \frac{\Delta \bx_t}{\frac{\partial L(\bx_t)}{\partial \bx}}\]
So we can use an estimation of $\frac{1}{\frac{\partial^2 L(\bx_t)}{\partial \bx_t^2}} $ to replace the computationally expensive $\bH^{-1}$:
\begin{equation}
\frac{\Delta \bx_t}{\frac{\partial L(\bx_t)}{\partial \bx_t}} \sim \frac{\rms[\Delta \bx]_{t-1}}{\rms[\bg]_t}.
\end{equation}
%where 
%
%\[\rms[X]_t = \sqrt{E[X^2]_t + \epsilon}, X = \Delta x, g \]
%
%\[ E[X^2]_t = \rho E[X^2]_{t-1} + (1 - \rho ) X_t^2, X = \Delta x, g\]
This is an approximation to the diagonal Hessian using only RMS measures of $\bg$ and $\Delta \bx$, and results in the update step whose units are matched:
\begin{equation}
	\Delta \bx_t = -\frac{\rms[\Delta \bx]_{t-1}}{\rms[\bg]_t} \odot \bg_t.
\end{equation}
The idea of AdaDelta from the second method overcomes the annoying choosing of learning rate. Similarly, in \citet{kingma2014adam}, an exponentially decaying average is incorporated into the gradient information such that convergence in online convex setting is improved. Meanwhile, the diminishing problem was also attacked in second order methods \citep{schaul2013no}. The idea of averaging gradient or its alternative is not unique, previously in \citet{moulines2011non}, Polyak-Ruppert averaging has been shown to improve the convergence of vanilla
SGD \citep{ruppert1988efficient, polyak1992acceleration}. 
%Apart from the methods discussed above, 
Other stochastic optimization methods, including the natural Newton method, AdaMax, Nadam, all set the step-size per-dimension by estimating curvature from first-order information \citep{le2010fast, kingma2014adam, dozat2016incorporating}. The LAMB adopts layerwise normalization due to layerwise adaptivity \citep{you2019large} and we shall not go into the details. 

%\subsection{RMSprop}
%It is identical to the first update version of AdaDelta that I derive above, i.e.:
%
%\[\Delta x_i^t=- \frac{\eta}{\rms[g_i]_t} g_{t,i}\]
%
%where 
%
%\[E[g_i^2]_t = \rho E[g_i^2]_{t-1} + (1 - \rho) g_{t,i}^2\]
%
%\[ \rms[g_i]_t = \sqrt{E[g_i^2]_t + \epsilon} \]
%
%RMSprop as well divides the learning rate by an exponentially decaying average of squared gradients. Hinton suggests $\rho$ to be set to 0.9, while a good default value for the learning rate $\eta$ is 0.001 .

\section{AdaSmooth Method}\label{section:adaer}
In this section we will discuss the effective ratio based on previous updates in the stochastic optimization process and how to apply it to accomplish adaptive learning rates per-dimension via the flexible smoothing constant, hence the name AdaSmooth.
The idea presented in the paper is derived from the first form of AdaDelta \citep{zeiler2012adadelta} in order to improve two main drawbacks of the method: 1) consider only the absolute value of the gradients rather than the total movement in each dimension; 2) the need for manually selected hyper-parameters. 
\subsection{Effective Ratio (ER)}
\citet{kaufman2013trading, kaufman1995smarter} suggested replacing the smoothing constant in the EMA formula with a constant based on the \textit{efficiency ratio} (ER). And the ER is shown to provide promising results for financial forecasting via classic quantitative strategies \cite{lu2022exploring} where the ER of the closing price is calculated to decide the trend of the asset. This indicator is designed to measure the \textit{strength of a trend}, defined within a range from -1.0 to +1.0 where the larger magnitude indicates a larger upward or downward trend.
Recently, \citet{lu2022reducing} shows the ER can be utilized to reduce overestimation and underestimation in time series forecasting.
Given the window size $M$ and a series $\{h_1, h_2, \ldots, h_T\}$, it is calculated with a simple formula:
\noindent
\begin{equation}
\begin{aligned}
	e_t  &= \frac{s_t}{n_t}= \frac{h_{t} - h_{t-M}}{\sum_{i=0}^{M-1} |h_{t-i} - h_{t-1-i}|}\\
&= \frac{\text{Total move for a period}}{\text{Sum of absolute move for each bar}},
\end{aligned}
\end{equation}
where $e_t$ is the ER of the series at time $t$. 
At a strong trend (i.e., the input series is moving in a certain direction, either up or down) the ER will tend to 1 in absolute value; if there is no directed movement, it will be a little more than 0. 

Instead of calculating the ER of the closing price of asset, we want to calculate the ER of the moving direction in the update methods for each parameter. And in the descent methods, we care more about how much each parameter moves apart from its initial point in each period, either move positively or negatively. So here we only consider the absolute value of the ER. To be specific, the ER for the parameters in the proposed method is calculated as follows:
\begin{equation}\label{eqution:signoiase-er-delta}
\begin{aligned}
\be_t  = \frac{\bs_t}{\bn_t}&= \frac{| \bx_t -  \bx_{t-M}|}{\sum_{i=0}^{M-1} | \bx_{t-i} -  \bx_{t-1-i}|}\\
&= \frac{| \sum_{i=0}^{M-1} \Delta \bx_{t-1-i}|}{\sum_{i=0}^{M-1} | \Delta \bx_{t-1-i}|},
\end{aligned}
\end{equation}
where $\be_t \in \real^d$ whose $i$-th element $e_{t,i}$ is in the range of $ [0, 1]$ for all $i$ in $[1,2,\ldots, d]$. A larger value of $e_{t,i}$ indicates the descent method in the $i$-th dimension is moving in a certain direction; while a smaller value approaching 0 means the parameter in the $i$-th dimension is moving in a zigzag pattern, interleaved by positive and negative movement. In practice, and in all of our experiments, the $M$ is selected to be the batch index for each epoch. That is, $M=1$ if the training is in the first batch of each epoch; and $M=M_{\text{max}}$ if the training is in the last batch of the epoch where $M_{\text{max}}$ is the maximal number of batches per epoch. In other words, $M$ ranges in $[1, M_{\text{max}}]$ for each epoch. Therefore, the value of $e_{t,i}$ indicates the movement of the $i$-th parameter in the most recent epoch. Or even more aggressively, the window can range from 0 to the total number of batches seen during the whole training progress. The adoption of the adaptive window size $M$ rather than a fixed one has a benefit that we do not need to keep the past $M+1$ steps $\{ \bx_{t-M},  \bx_{t-M+1}, \ldots,  \bx_t\}$ to calculate the signal and noise vectors $\{\bs_t,\bn_t\}$ in Eq~\eqref{eqution:signoiase-er-delta} since they can be obtained in an accumulated fashion.

\subsection{AdaSmooth}\label{section:adaer-after-er}
\begin{figure}%[H]
	\centering
	\includegraphics[width=0.47\textwidth]{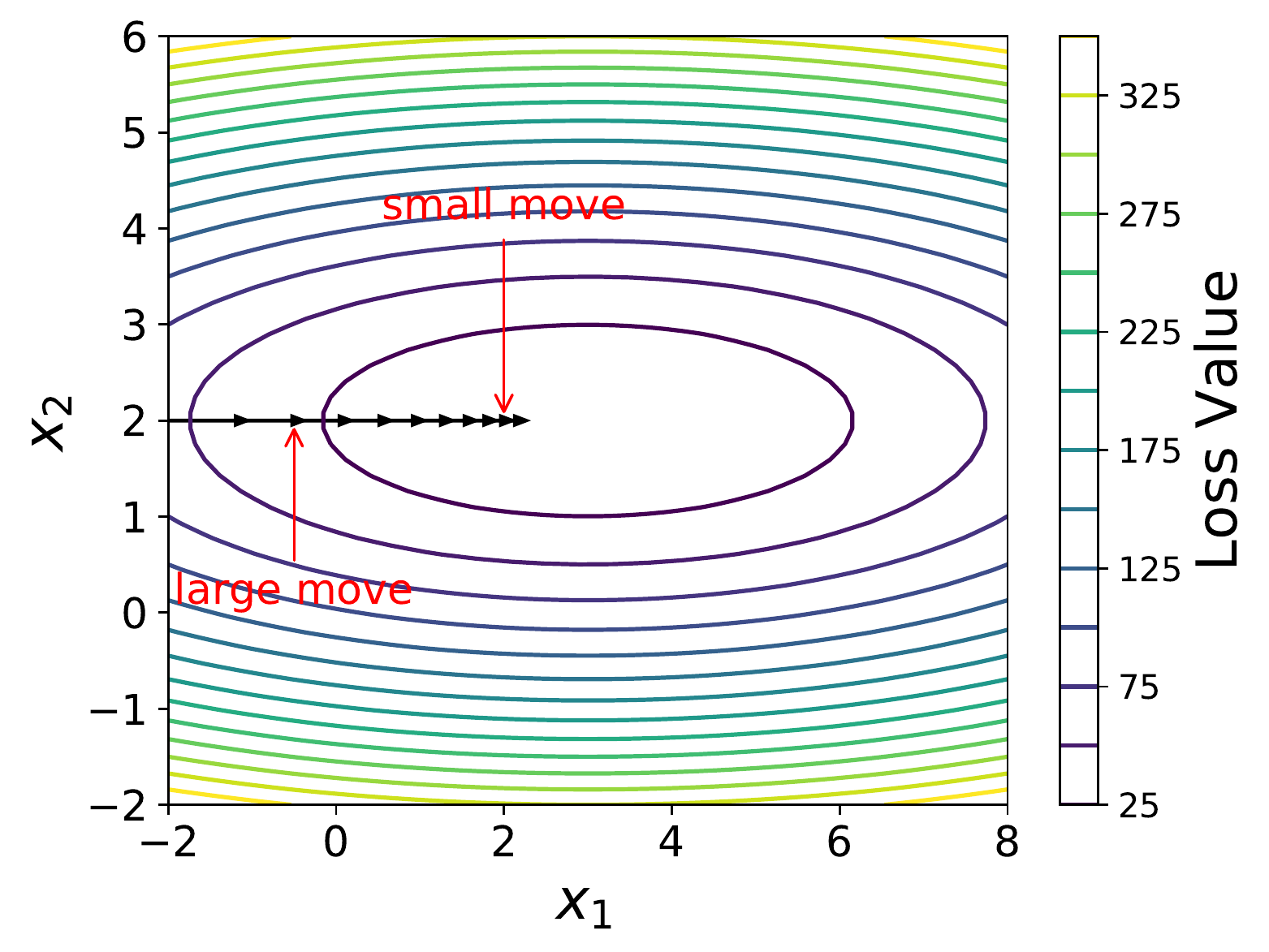}
	\caption{Demonstration of how the effective ratio works. Stochastic optimization tends to move a large step when it is far from the (local) minima; and a relatively small step when it is close to the (local) minima.}
	\label{fig:er-explain}
\end{figure}

If the ER in magnitude of each parameter is small (approaching 0), the movement in this dimension is zigzag, the proposed AdaSmooth method tends to use a long period average as the scaling constant to slow down the movement in that dimension. When the absolute ER per-dimension is large (tend to 1), the path in that dimension is moving in a certain direction (not zigzag), and the learning actually is happening and the descent is moving in a correct direction where the learning rate should be assigned to a relatively large value for that dimension. Thus the AdaSmooth tends to choose a small period which leads to a small compensation in the denominator; since the gradients in the closer periods are small when it's near the (local) minima. A particular example is shown in Figure~\ref{fig:er-explain}, where the descent is moving in a certain direction, and the gradient in the near periods is small; if we choose a larger period to compensate for the denominator, the descent will be slower due to the large factored denominator.
In short, we want a smaller period to calculate the exponential average of the squared gradients in Eq~\eqref{equation:adagradwin} if the update is moving in a certain direction without a zigzag pattern; while when the parameter is updated in a zigzag basis, the period for the exponential average should be larger.

The obtained value of ER is used in the exponential smoothing formula.  Now, what we want to go further is to set the time period $N$ discussed in Eq~\eqref{equation:ema_smooting_constant} to be a smaller value when the ER tends to 1 in absolute value; or a larger value when the ER moves towards 0. When $N$ is small, $\text{SC} $ is known as a ``\textit{fast SC}"; otherwise, $\text{SC} $ is known as a ``\textit{slow SC}". 

For example, let the small time period be $N_1=3$, and the large time period be $N_2=199$.
The smoothing ratio for the fast movement must be as for EMA with period $N_1$ (``fast SC" = $\frac{2}{N_1+1}$ = 0.5), and for the period of no trend EMA period must be equal to $N_2$ (``slow SC" = $\frac{2}{N_2+1}$ = 0.01). 
Thus the new changing smoothing constant is introduced, called the ``\textit{scaled smoothing constant}" (SSC), denoted by a vector $\bc_t\in \real^d$:
\begin{equation}
	\bc_t =  ( \text{fast SC} - \text{slow SC}) \times \be_t   + \text{slow SC}.
\end{equation}
By Eq~\eqref{equation:ema_smooting_constant}, we can define the \textit{fast decay constant} $\rho_1=1-\frac{2}{N_1+1}$, and the \textit{slow decay constant} $\rho_2 = 1-\frac{2}{N_2+1}$. Then the scaled smoothing constant vector can be obtained by:
\begin{equation}
	\bc_t =  ( \rho_2- \rho_1) \times \be_t   + (1-\rho_2),
\end{equation}
where the smaller $\be_t$, the smaller $\bc_t$.
For a more efficient influence of the obtained smoothing constant on the averaging period, Kaufman recommended squaring it.
The final calculation formula then follows:
\begin{equation}\label{equation:squared-ssc}
E[\bg^2]_t  =  \bc_t^2 \odot \bg_{t}^2  +  \left(1-\bc_t^2 \right)\odot E[\bg^2]_{t-1}.
\end{equation}
%or after rearrangement:
%\begin{equation}
%E[\bg^2]_t = E[\bg^2]_{t-1}+ \bc_t^2 \odot (\bg_{t}^2 -  E[\bg^2]_{t-1}).
%\end{equation}
We notice that $N_1=3$ is a small period to calculate the average (i.e., $\rho_1=1-\frac{2}{N_1+1}=0.5$) such that the EMA sequence will be noisy if $N_1$ is less than 3. Therefore, the minimal value of $\rho_1$ in practice is set to be larger than 0.5 by default. While $N_2=199$ is a large period to compute the average (i.e., $\rho_2=1-\frac{2}{N_2+1}=0.99$) such that the EMA sequence almost depends only on the previous value leading to the default value of $\rho_2$ no larger than 0.99. Experiment study will show that the AdaSmooth update will be insensitive to the hyper-parameters in the sequel. We also carefully notice that when $\rho_1=\rho_2$, the AdaSmooth algorithm recovers to the RMSProp algorithm with decay constant $\rho=1-(1-\rho_2)^2$ since we square it in Eq~\eqref{equation:squared-ssc}. After developing the AdaSmooth method, we realize the main idea behind it is similar to that of SGD with Momentum: to speed up (compensate less in the denominator) the learning along dimensions where the gradient consistently points in the same direction; and to slow the pace (compensate more in the denominator) along dimensions in which the sign of the gradient continues to change.

%After developing the AdaSmooth and discussing the empirical results, we find the AdaSmooth is very similar to the global learning rate annealing in Section~\ref{section:learning-rate-annealing}. However, the AdaSmooth can anneal the learning rate per-dimension.

Empirical evidence shows the ER used in simple moving average with a fixed windows size $w$ can also reflect the trend of the series/movement in quantitative strategies \citep{lu2022exploring}. However, this again needs to store $w$ previous squared gradients in the AdaSmooth case, making it inefficient and we shall not adopt this extension.

\subsection{AdaSmoothDelta}\label{section:adasmoothdelta}
Notice the ER can also be applied to the AdaDelta setting:
\begin{equation}\label{equation:adasmoothdelta}
	\Delta \bx_t = -\frac{\sqrt{E[\Delta \bx^2]_t}}{\sqrt{E[\bg^2]_t+\epsilon}} \odot \bg_t,
\end{equation}
where 
\begin{equation}\label{equation:adasmoothdelta111}
E[\bg^2]_t  =  \bc_t^2 \odot \bg_{t}^2  +  \left(1-\bc_t^2 \right)\odot E[\bg^2]_{t-1} ,
\end{equation}
and 
\begin{equation}\label{equation:adasmoothdelta222}
	E[\Delta \bx^2]_t = (1-\bc_t^2) \odot \Delta \bx^2_t+ \bc_t^2 \odot E[\Delta \bx^2]_{t-1},
\end{equation}
in which case the difference in $E[\Delta \bx^2]_t$ is to choose a larger period when the ER is small. This is reasonable in the sense that $E[\Delta \bx^2]_t$ appears in the numerator while $E[\bg^2]_t$ is in the denominator of Eq~\eqref{equation:adasmoothdelta} making their compensation towards different directions. Or even, a fixed decay constant can be applied for $E[\Delta \bx^2]_t$:
\begin{equation}
	E[\Delta \bx^2]_t = (1-\rho_2)  \Delta \bx^2_t+ \rho_2  E[\Delta \bx^2]_{t-1},
\end{equation}
The AdaSmoothDelta optimizer introduced above further alleviates the need for a hand specified global learning rate which is set to $\eta=1$ from the Hessian context. However, due to the adaptive smoothing constants in Eq~\eqref{equation:adasmoothdelta111} and \eqref{equation:adasmoothdelta222}, the $E[\bg^2]_t $ and $E[\Delta \bx^2]_t$ are less locally smooth making it less insensitive to the global learning rate than the AdaDelta method. Therefore, a smaller global learning rate, e.g., $\eta=0.5$ is favored in AdaSmoothDelta. The full procedure for computing AdaSmooth is then formulated in Algorithm~\ref{algo:adasmooth}.

\begin{algorithm}[tb]
\caption{Computing AdaSmooth at iteration $t$: the proposed AdaSmooth algorithm. All operations on vectors are element-wise. Good default settings for the tested tasks are $\rho_1=0.5, \rho_2=0.99, \epsilon=1e-6, \eta=0.001$; see Section~\ref{section:adaer-after-er} or Eq~\eqref{equation:ema_smooting_constant} for a detailed discussion on the explanation of the decay constants' default values. Empirical study in Section~\ref{section:ader_experiments} shows that the AdaSmooth algorithm is not sensitive to the hypermarater $\rho_2$, while $\rho_1=0.5$ is relatively a lower bound in this setting. The AdaSmoothDelta iteration can be calculated in a similar way.}
\label{alg:computer-adaer}
\begin{algorithmic}[1]
\STATE {\bfseries Input:} initial parameter $\bx_1$, Constant $\epsilon$;
\STATE {\bfseries Input:} global learning rate $\eta$, by default $\eta=0.001$;
\STATE {\bfseries Input:} fast decay constant $\rho_1$, slow decay constant $\rho_2$;
\STATE {\bfseries Input:} assert $\rho_2>\rho_1$, by default $\rho_1=0.5$, $\rho_2=0.99$;
\FOR{$t=1:T$ } 
\STATE Compute gradient $\bg_t = \nabla L(\bx_t)$;
\STATE Compute ER $\be_t=\frac{| \bx_t -  \bx_{t-M}|}{\sum_{i=0}^{M-1} | \Delta \bx_{t-1-i}|}$ ;
\STATE Compute smoothing $\bc_t =  ( \rho_2- \rho_1) \times \be_t   + (1-\rho_2)$;
\STATE Compute normalization term: $$E[\bg^2]_t  =  \bc_t^2 \odot \bg_{t}^2  +  \left(1-\bc_t^2 \right)\odot E[\bg^2]_{t-1} ;$$
\STATE Compute step $\Delta \bx_t =- \frac{\eta}{\sqrt{E[\bg^2]_t+\epsilon}}  \odot \bg_{t}$;
\STATE Apply update $\bx_t = \bx_{t-1} + \Delta \bx_t$;
\ENDFOR
\STATE {\bfseries Return:} resulting parameters $\bx_t$, and the loss $L(\bx_t)$.
\end{algorithmic}\label{algo:adasmooth}
\end{algorithm}

\begin{figure}[!h]
\centering
\subfigure[MNIST training Loss]{\includegraphics[width=0.491\textwidth, ]{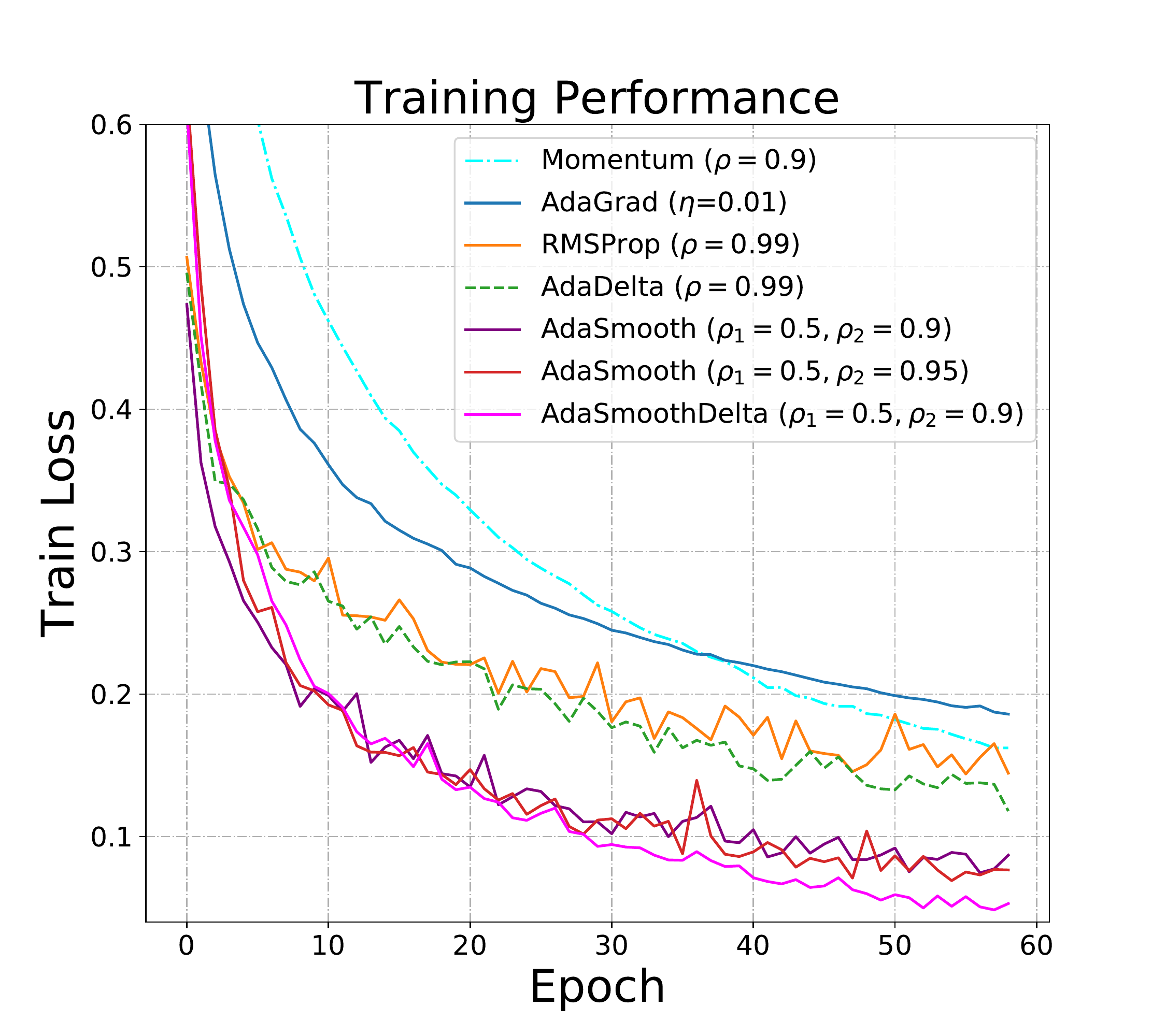} \label{fig:mnist_loss_train_mlp}}
\subfigure[Census Income training loss]{\includegraphics[width=0.491\textwidth]{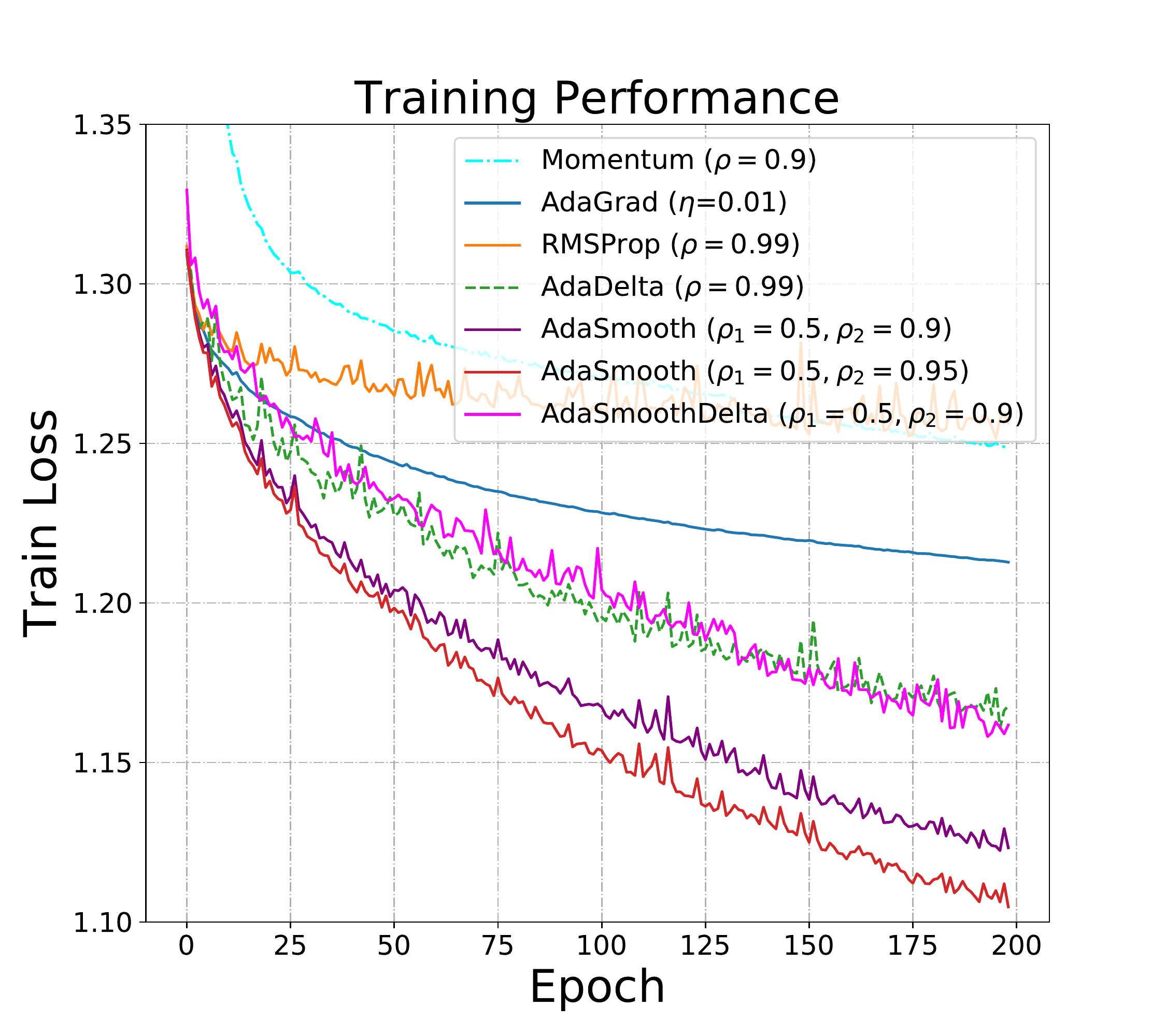} \label{fig:census_loss_train_mlp}}
\caption{\textbf{MLP:} Comparison of descent methods on MNIST digit and Census Income data sets for 60 and 200 epochs with MLP.}
\label{fig:mnist_census_loss_MLP}
\end{figure}

\section{Experiments}\label{section:ader_experiments}
To evaluate the strategy and demonstrate the main advantages of the proposed AdaSmooth method, 
we conduct experiments with different machine learning models; and different data sets including 
real handwritten digit classification task, MNIST \citep{lecun1998mnist} \footnote{It has a training set of 60,000 examples, and a test set of 10,000 examples.}, and Census Income \footnote{Census income data has 48842 number of samples where 70\% of them are used as training set in our case: https://archive.ics.uci.edu/ml/datasets/Census+Income.} data sets are used.
In all scenarios, same parameter initialization is adopted when training with different stochastic optimization algorithms. We compare the results in terms of convergence speed and generalization. In a wide range of scenarios across various models, AdaSmooth improves optimization rates, and leads to out-of-sample performances that are as good or better than existing stochastic optimization algorithms in terms of loss and accuracy.

\subsection{Experiment: Multi-Layer Perceptron}
\begin{table}[!h]
	\begin{tabular}{lll}
		\hline
		Method & MNIST &  Census \\ \hline
		Momentum ($\rho=0.9$) & 98.64\% & 85.65\%\\
		AdaGrad ($\eta$=0.01) & 98.55\%& 86.02\%\\
		RMSProp ($\rho=0.99$) & 99.15\%& 85.90\%\\
		AdaDelta ($\rho=0.99$) & 99.15\%& 86.89\%\\
		AdaSmooth ($\rho_1=0.5, \rho_2=0.9$) & \textbf{99.34}\%& \textbf{86.94}\%\\
		AdaSmooth ($\rho_1=0.5, \rho_2=0.95$) & \textbf{99.45}\%& \textbf{87.10}\%\\
		AdaSmDel. ($\rho_1=0.5, \rho_2=0.9$) & \textbf{99.60}\%& {86.86}\%\\
		\hline
	\end{tabular}
	\caption{\textbf{MLP}: Best in-sample evaluation in training accuracy (\%). AdaSmDel is short for AdaSmoothDelta.}
	\label{fig:mlp_table_perform}
\end{table}

\begin{table}[!h]
	\begin{tabular}{lll}
		\hline
		Method & MNIST &  Census \\ \hline
		Momentum ($\rho=0.9$) & 94.38\%& 83.13\%\\
		AdaGrad ($\eta$=0.01) & 96.21\%& 84.40\%\\
		RMSProp ($\rho=0.99$) & 97.14\%& 84.43\%\\
		AdaDelta ($\rho=0.99$) & 97.06\%&84.41\%\\
		AdaSmooth ($\rho_1=0.5, \rho_2=0.9$) & 97.26\%& 84.46\%\\
		AdaSmooth ($\rho_1=0.5, \rho_2=0.95$) & 97.34\%& 84.48\%\\
		AdaSmDel. ($\rho_1=0.5, \rho_2=0.9$) & 97.24\% & \textbf{84.51}\%\\
		\hline
	\end{tabular}
	\caption{\textbf{MLP}: Best out-of-sample evaluation in test accuracy for the first 5 epochs. AdaSmDel is short for AdaSmoothDelta.}
	\label{fig:mlp_table_perform-test}
\end{table}

Multi-layer perceptrons (MLP, a.k.a., multi-layer neural networks) are powerful tools for solving machine learning tasks finding internal linear and nonlinear features behind the model inputs and outputs. We adopt the simplest MLP structure: an input layer, a hidden layer, and an output layer. We notice that rectified linear unit (Relu) outperforms Tanh, Sigmoid, and other nonlinear units in practice making it the default nonlinear function in our structures. Since dropout has become a core tool in training neural networks \citep{srivastava2014dropout}, we adopt 50\% dropout noise to the network architecture during training to prevent overfitting.   
To be more concrete, the detailed architecture for each fully connected layer is described by F$(\langle \textit{num outputs} \rangle:\langle \textit{activation function} \rangle)$; and for a dropout layer is described by
DP$(\langle \textit{rate} \rangle)$. Then the network structure we use can be described as follows:
\begin{equation}
	\text{F(128:Relu)DP(0.5)F(\text{num of classes}:Softmax)}.
\end{equation}
All methods are trained on mini-batches of 64 images per batch for 60 or 200 epochs through the training set. Setting the hyper-parameter to $\epsilon=1e-6$.
If not especially mentioned, the global learning rates are set to $\eta=0.001$ in all scenarios. While a relatively large learning rate ($\eta=0.01$) is used for AdaGrad method since its accumulated decaying effect; learning rate for the AdaDelta method is set to 1 as suggested by \citet{zeiler2012adadelta} and for the AdaSmoothDelta method is set to 0.5 as discussed in Section~\ref{section:adasmoothdelta} while we will show in the sequel the learning rate for AdaSmoothDelta will not influence the result significantly.
In Figure~\ref{fig:mnist_loss_train_mlp} and ~\ref{fig:census_loss_train_mlp} we compare SGD with Momentum, AdaGrad, RMSProp, AdaDelta, AdaSmooth and AdaSmoothDelta in optimizing the training set losses for MNIST and Census Income data sets respectively. The SGD with Momentum method does the worst in this case. AdaSmooth performs slightly better than AdaGrad and RMSProp in the MNIST case and much better than the latters in the Census Income case. AdaSmooth shows fast convergence from the initial epochs while continuing to reduce the training losses in both the two experiments. We here show two sets of slow decay constant for AdaSmooth, i.e., ``$\rho_2=0.9$" and ``$\rho_2=0.95$". Since we square the scaled smoothing constant in Eq~\eqref{equation:squared-ssc}, when $\rho_1=\rho_2=0.9$, the AdaSmooth recovers to RMSProp with $\rho=0.99$ (so as the AdaSmoothDelta and AdaDelta case). In all cases, the AdaSmooth results perform better while there is almost no difference between the results of AdaSmooth with various hyper-parameters in the MLP model. Table~\ref{fig:mlp_table_perform} shows the best training set accuracy for different algorithms, indicating the superiority of AdaSmooth. While we notice the best test set accuracy for various algorithms are very close; we only report the best ones for the first 5 epochs in Table~\ref{fig:mlp_table_perform-test}. In all scenarios, the AdaSmooth method converges slightly faster than other optimization methods in terms of the test accuracy.

\begin{figure*}[!h]
	\centering
	\subfigure[Train Loss]{\includegraphics[width=0.323\textwidth, ]{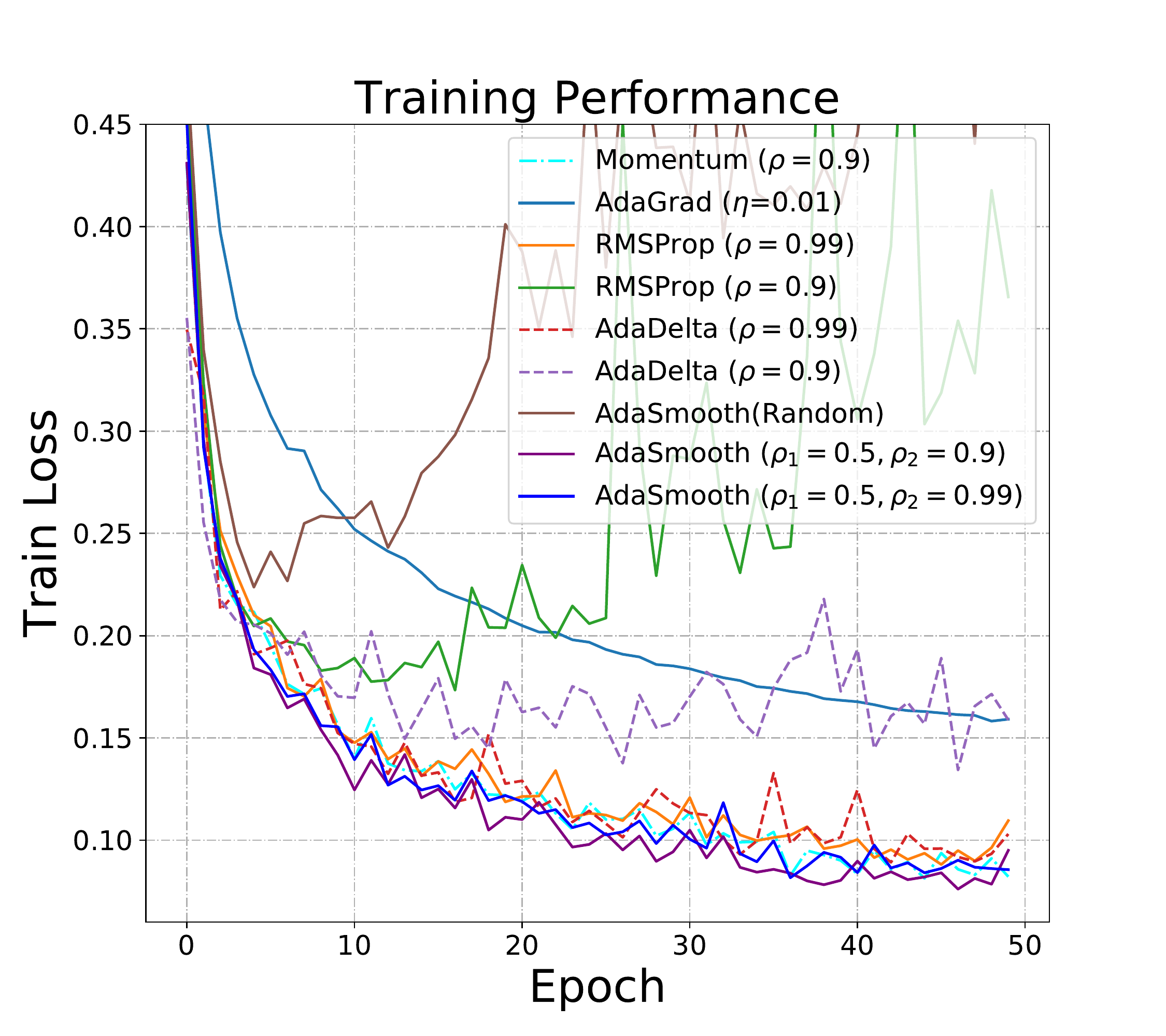} \label{fig:mnist_loss_train}}
	\subfigure[Test Loss]{\includegraphics[width=0.323\textwidth, ]{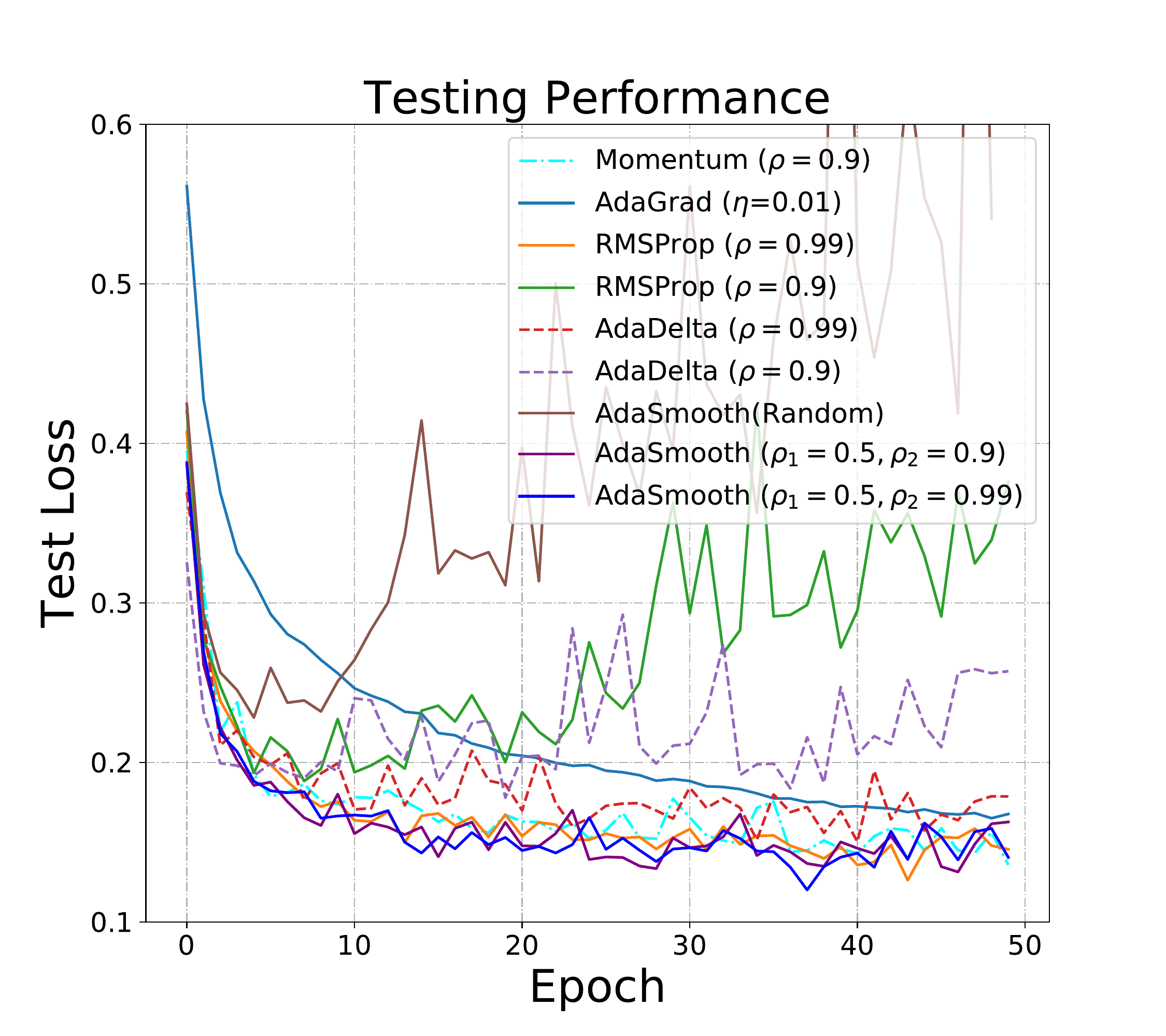} \label{fig:mnist_loss_test}}
	\subfigure[Test Accuracy]{\includegraphics[width=0.323\textwidth]{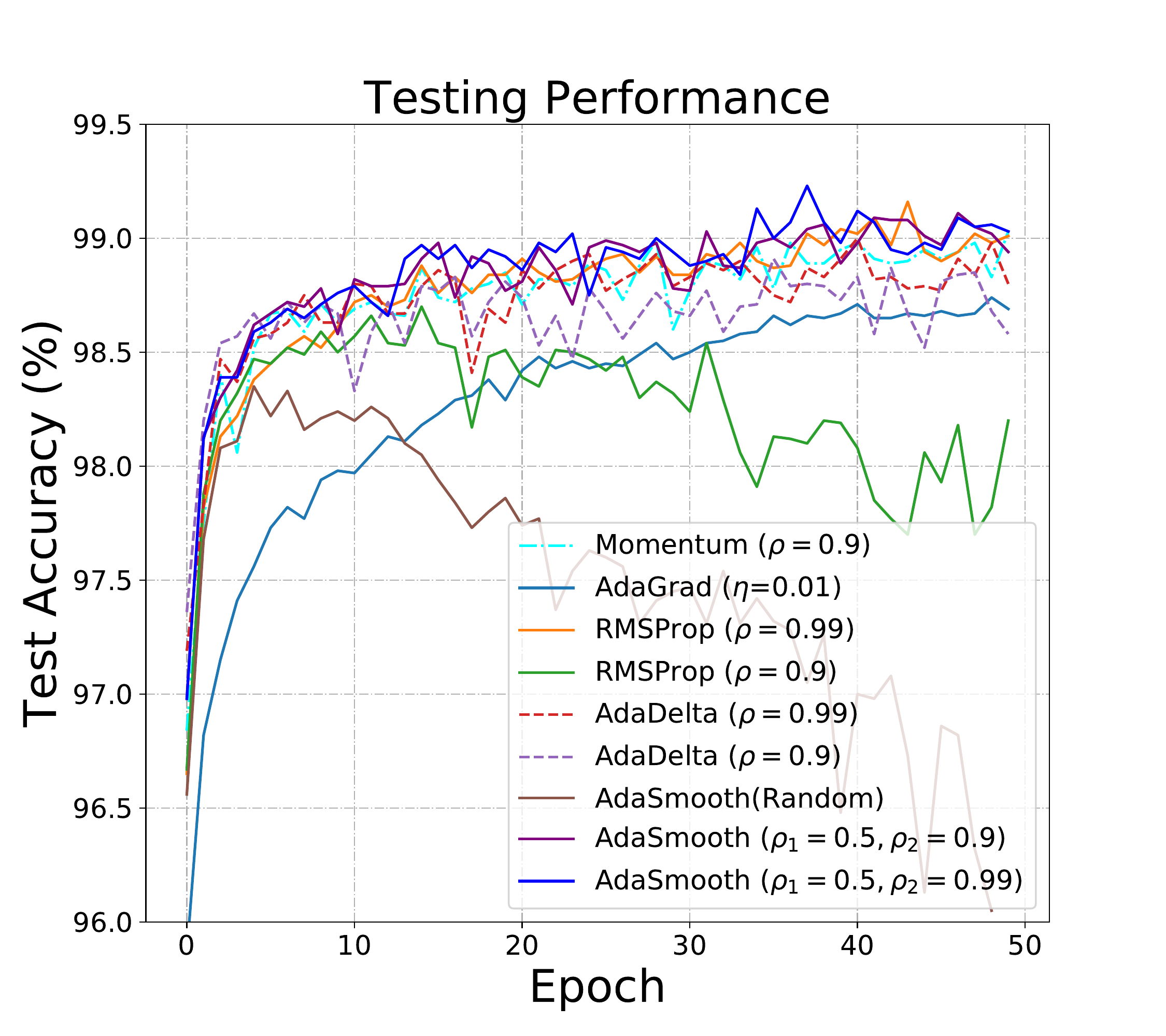} \label{fig:mnist_acc_test}}
	\caption{\textbf{CNN:} Comparison of descent methods on MNIST digit data set for 50 epochs with CNN. Though not significantly better than RMSProp or SGD with Momentum, AdaSmooth converges slightly faster and obtains better out-of-sample accuracy.}
	\label{fig:mnist_census_logisticregre}
\end{figure*}

\begin{figure}[!h]
	\centering
	\subfigure[Epoch 1]{\includegraphics[width=0.21\textwidth, ]{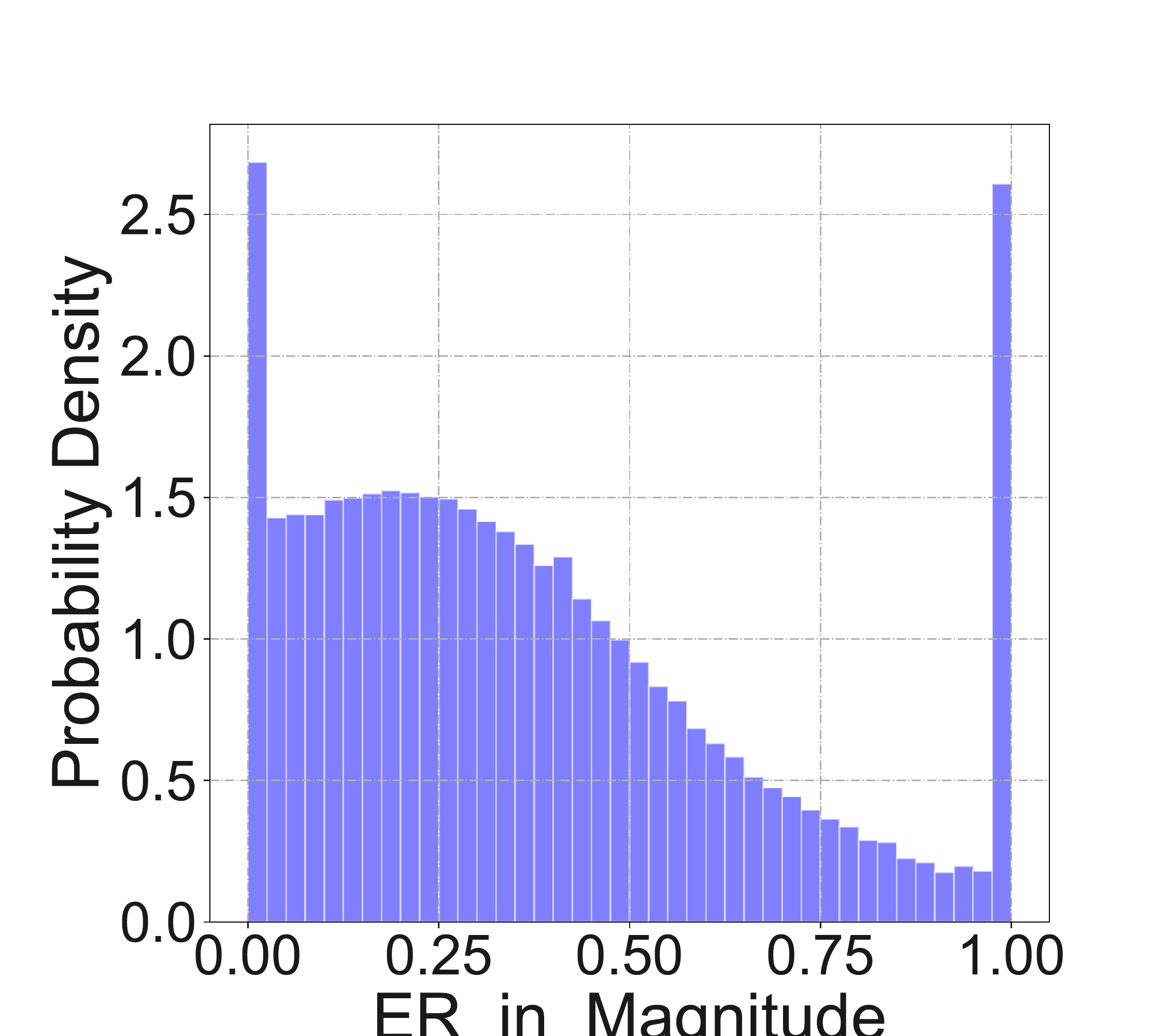} \label{fig:mnist_er1}}
	\subfigure[Epoch 5]{\includegraphics[width=0.21\textwidth]{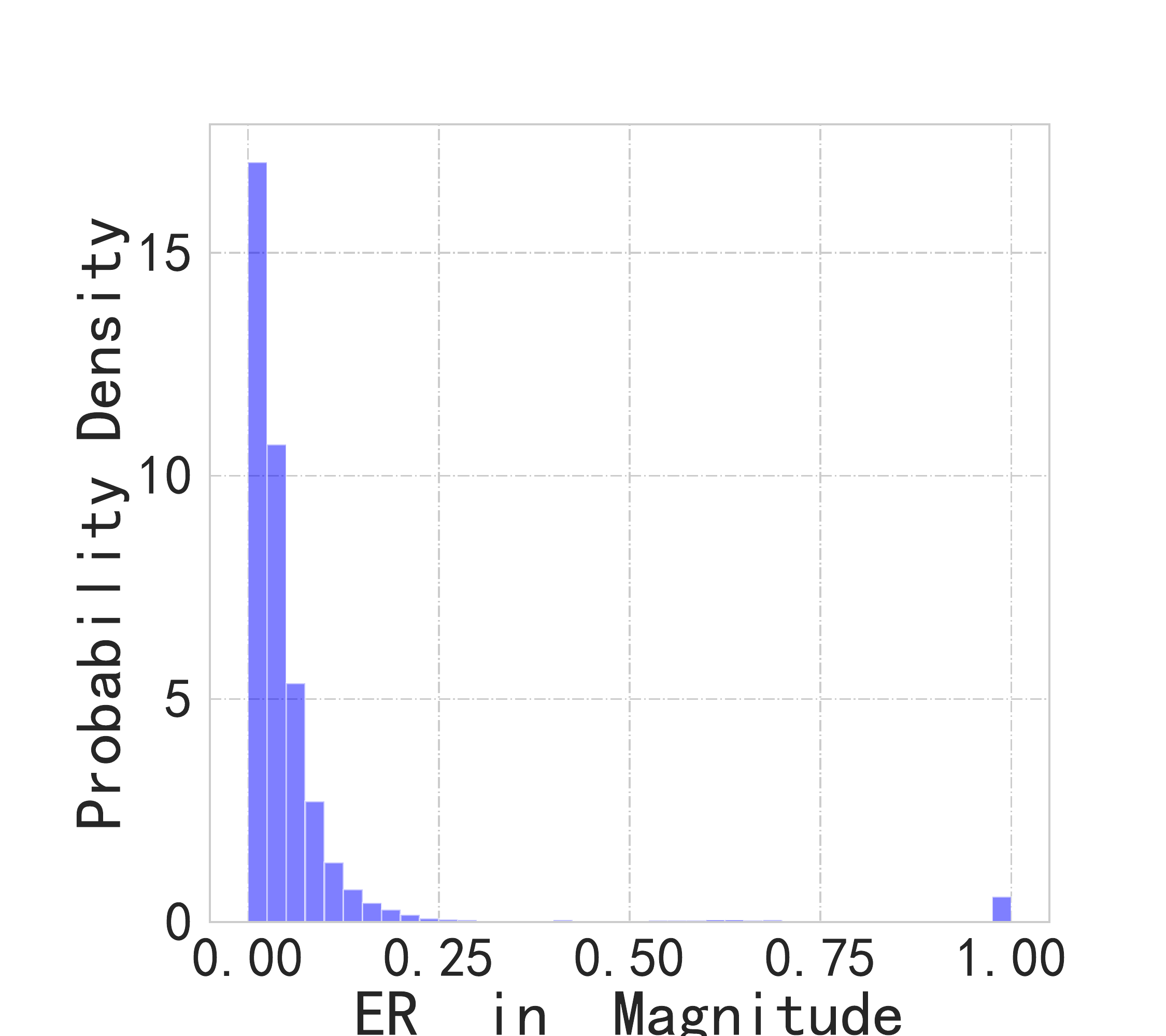} \label{fig:mnist_er2}}
	\subfigure[Epoch 20]{\includegraphics[width=0.21\textwidth]{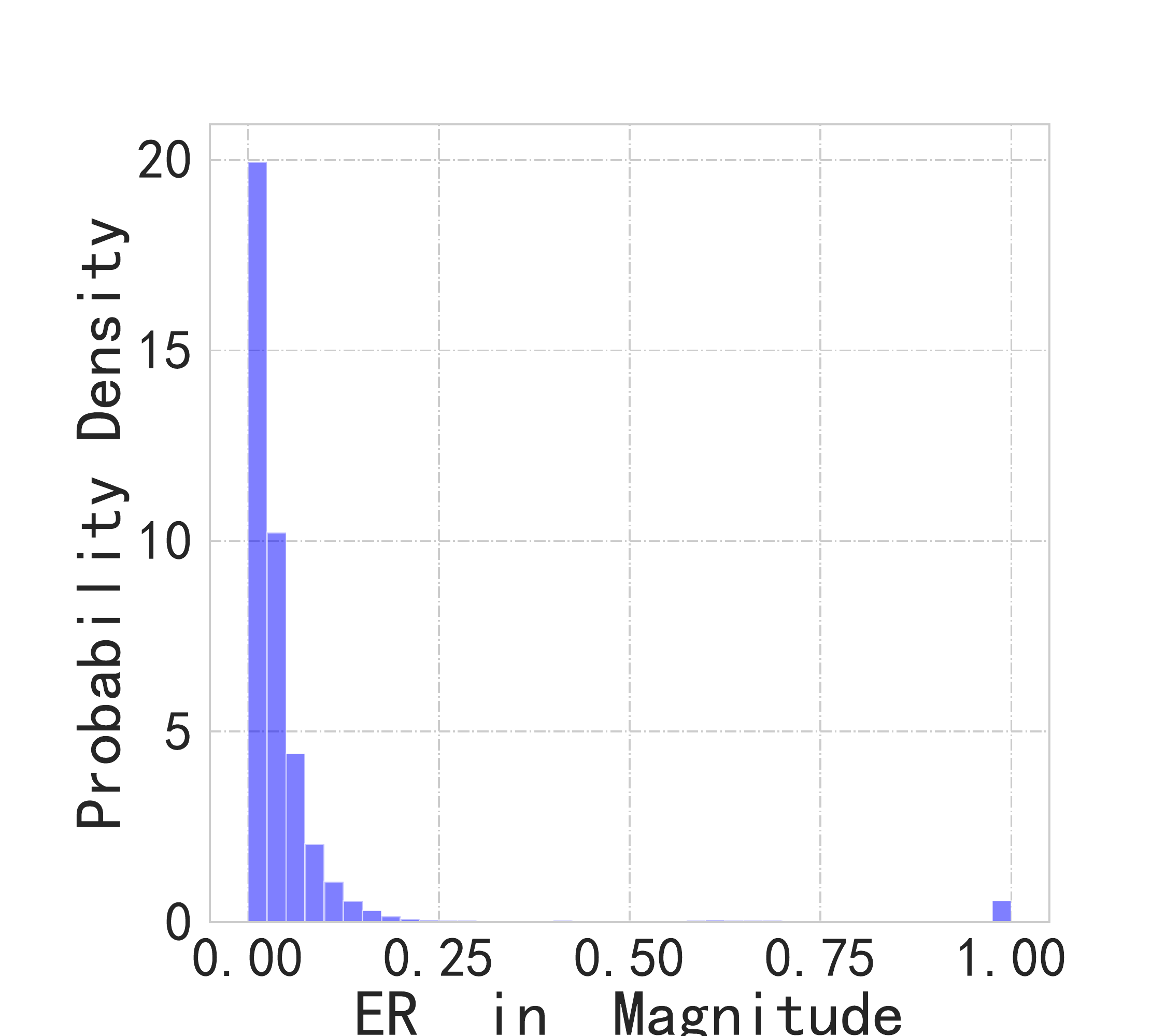} \label{fig:mnist_er3}}
	\subfigure[Epoch 50]{\includegraphics[width=0.21\textwidth]{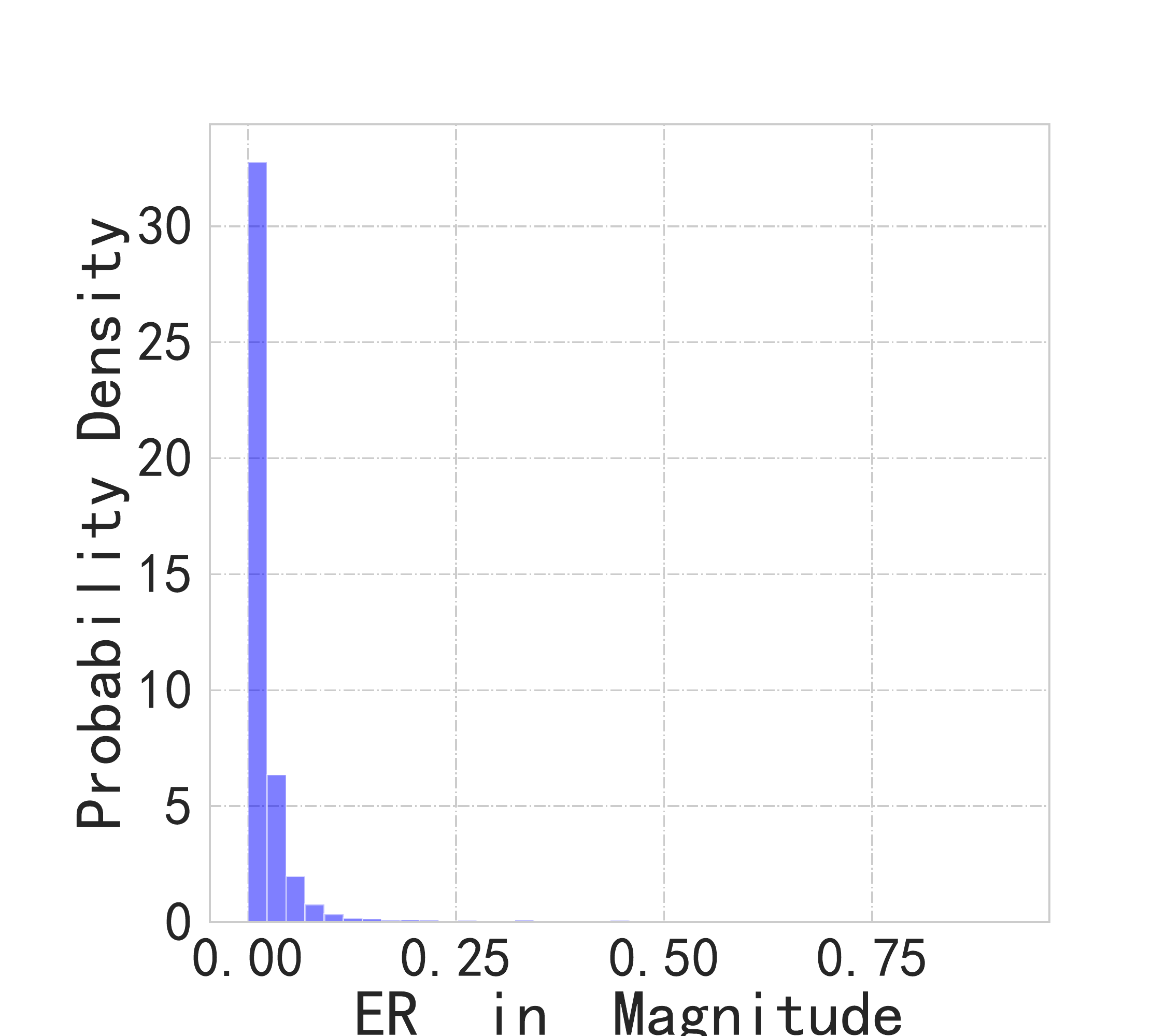} \label{fig:mnist_er4}}
	\caption{\textbf{CNN:} The distribution of ERs for all parameters in different epochs when training MNIST data for AdaSmooth ($\rho_1=0.5, \rho_2=0.99$) model. }
	\label{fig:mnist_er_dist}
\end{figure}
\begin{figure}[!h]
	\centering
	\subfigure[Threshold=0.02]{\includegraphics[width=0.21\textwidth, ]{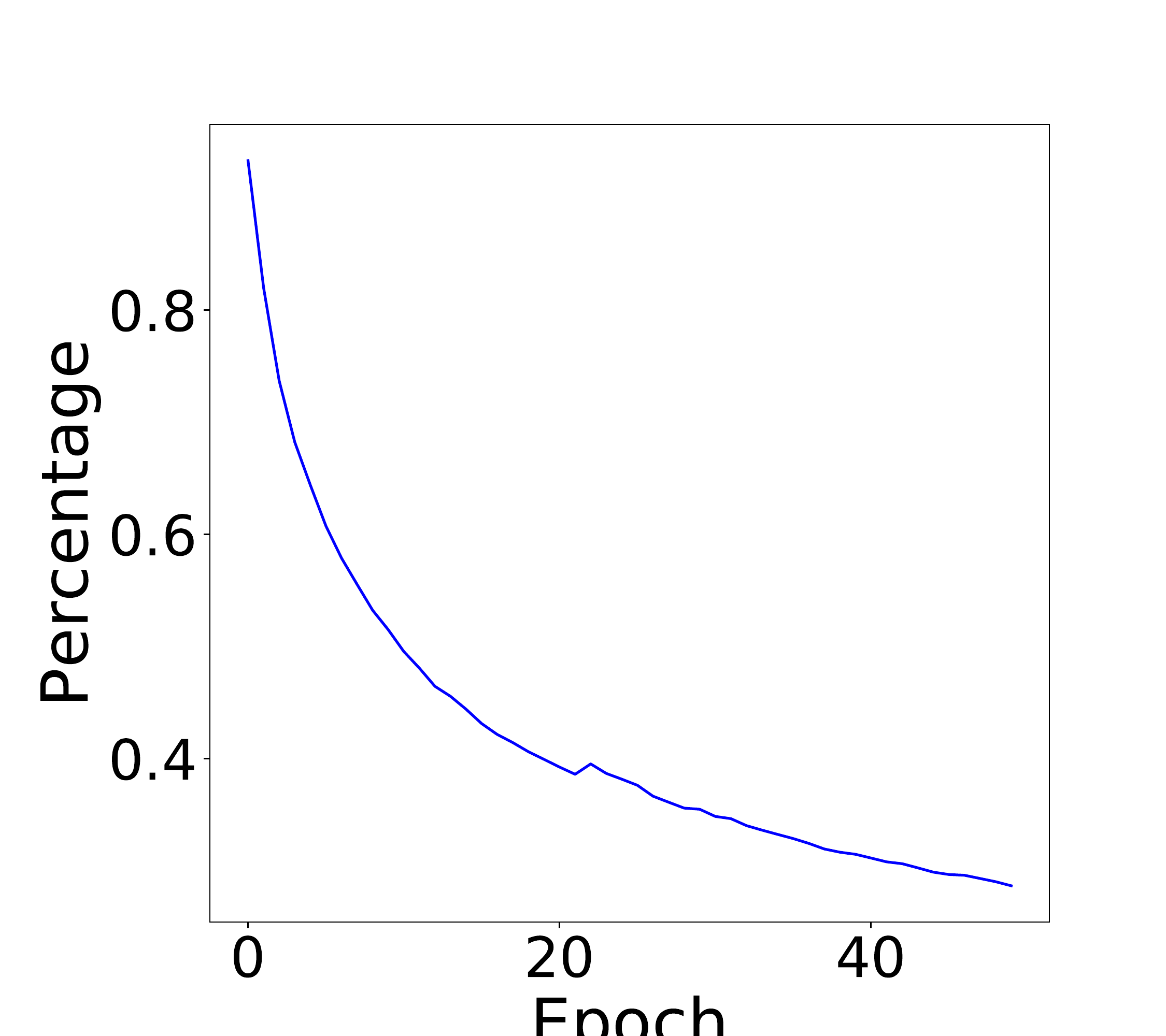} \label{fig:mnist_er_dist_thres1}}
	\subfigure[Threshold=0.04]{\includegraphics[width=0.21\textwidth]{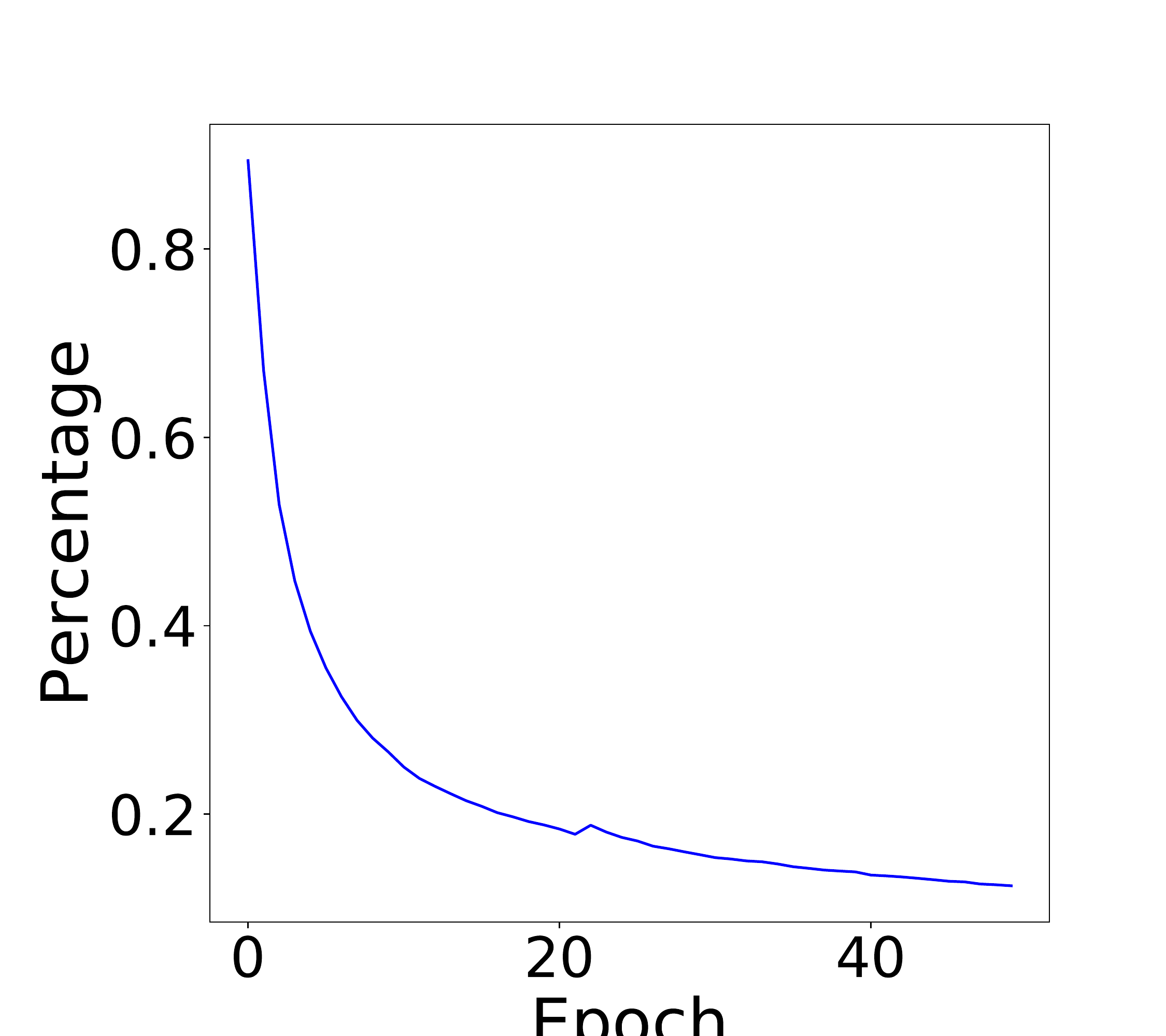} \label{fig:mnist_er_dist_thres2}}
	\subfigure[Threshold=0.06]{\includegraphics[width=0.21\textwidth]{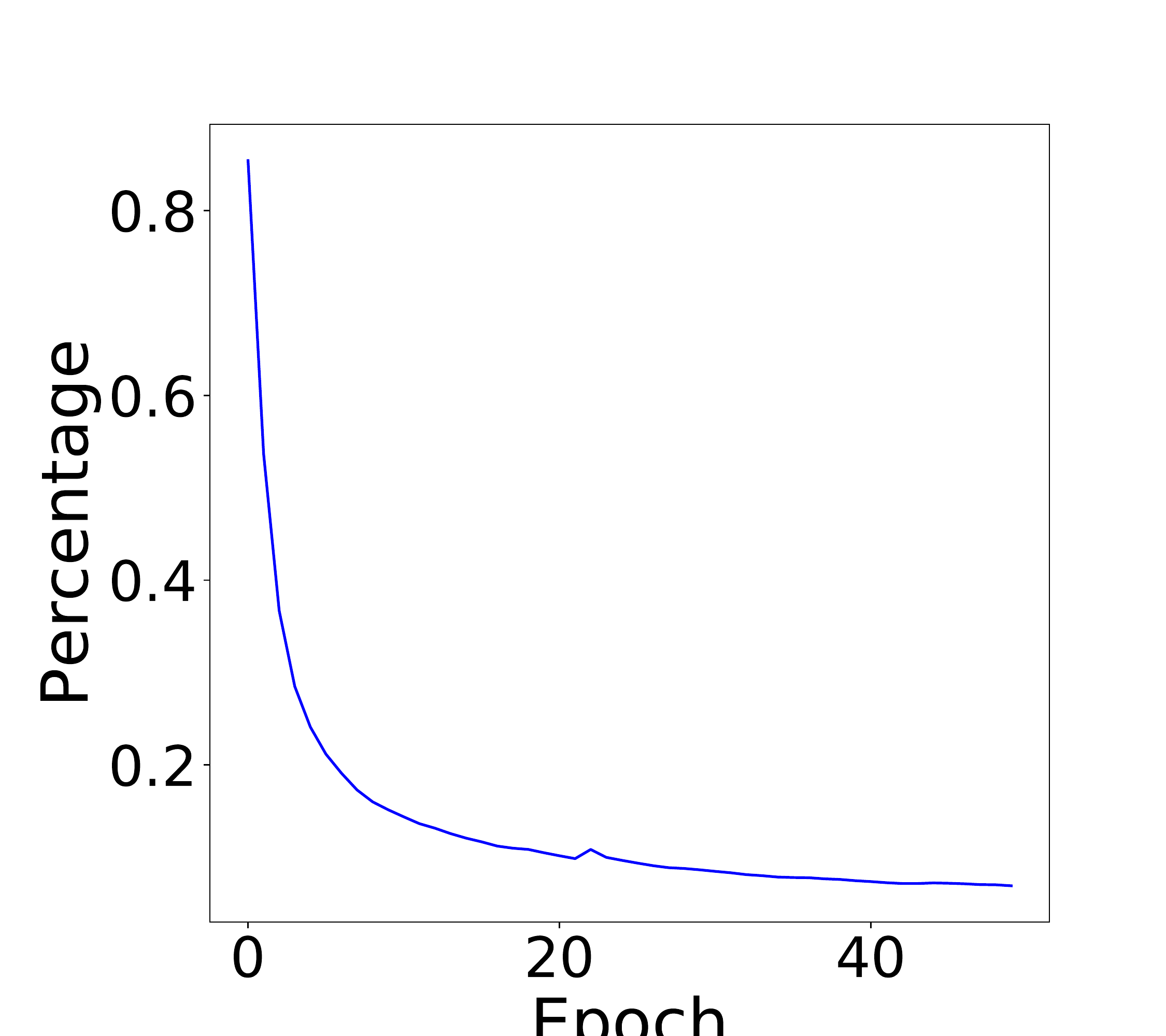} \label{fig:mnist_er_dist_thres3}}
	\subfigure[Threshold=0.1]{\includegraphics[width=0.21\textwidth]{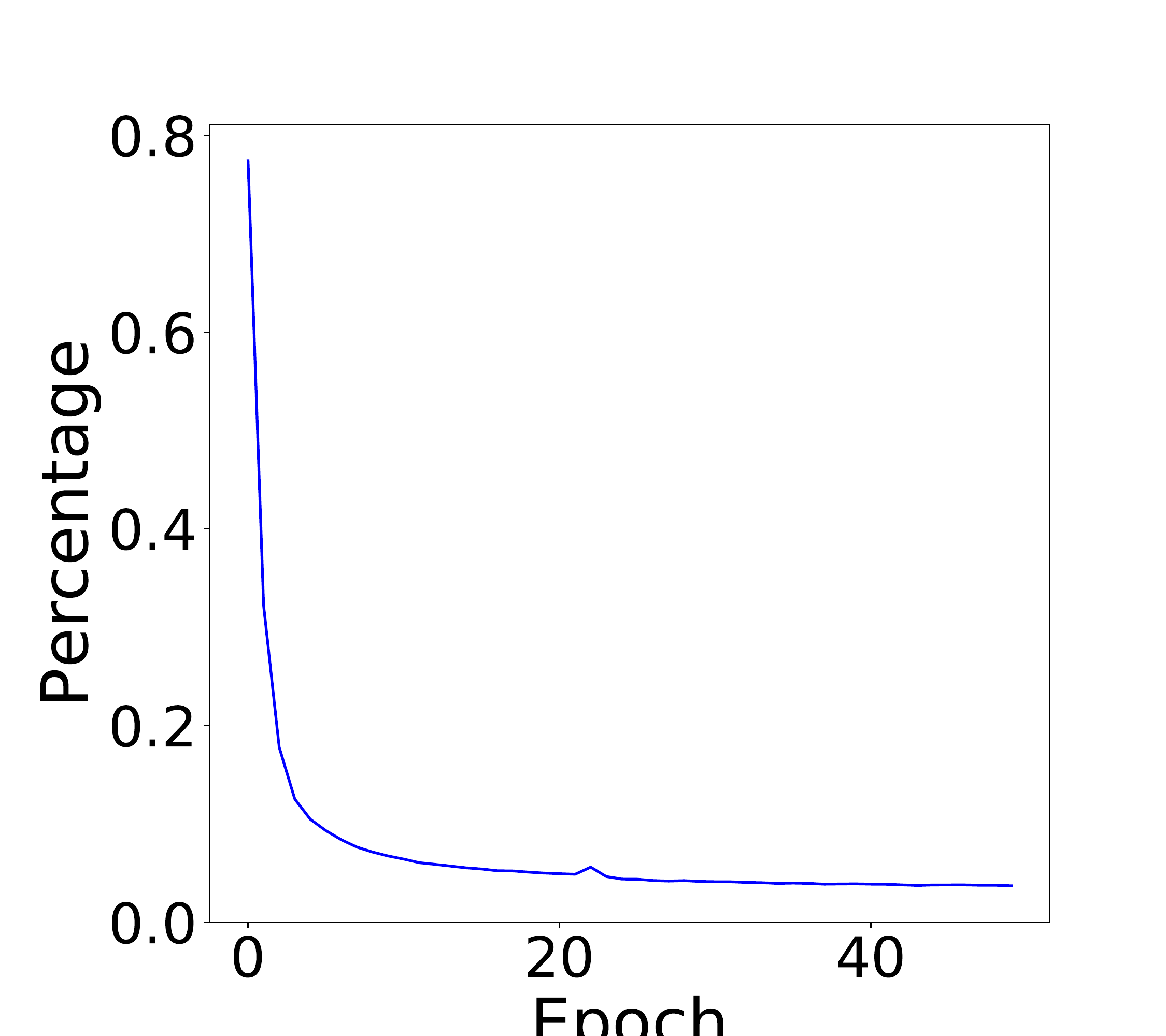} \label{fig:mnist_er_dist_thres4}}
	\caption{\textbf{CNN:} Percentage of ERs when they are larger than a threshold for different epochs in AdaSmooth ($\rho_1=0.5, \rho_2=0.99$) model.}
	\label{fig:mnist_er_dist_thres}
\end{figure}

\begin{figure}[!h]
	\centering
	\subfigure[Epoch 1]{\includegraphics[width=0.21\textwidth, ]{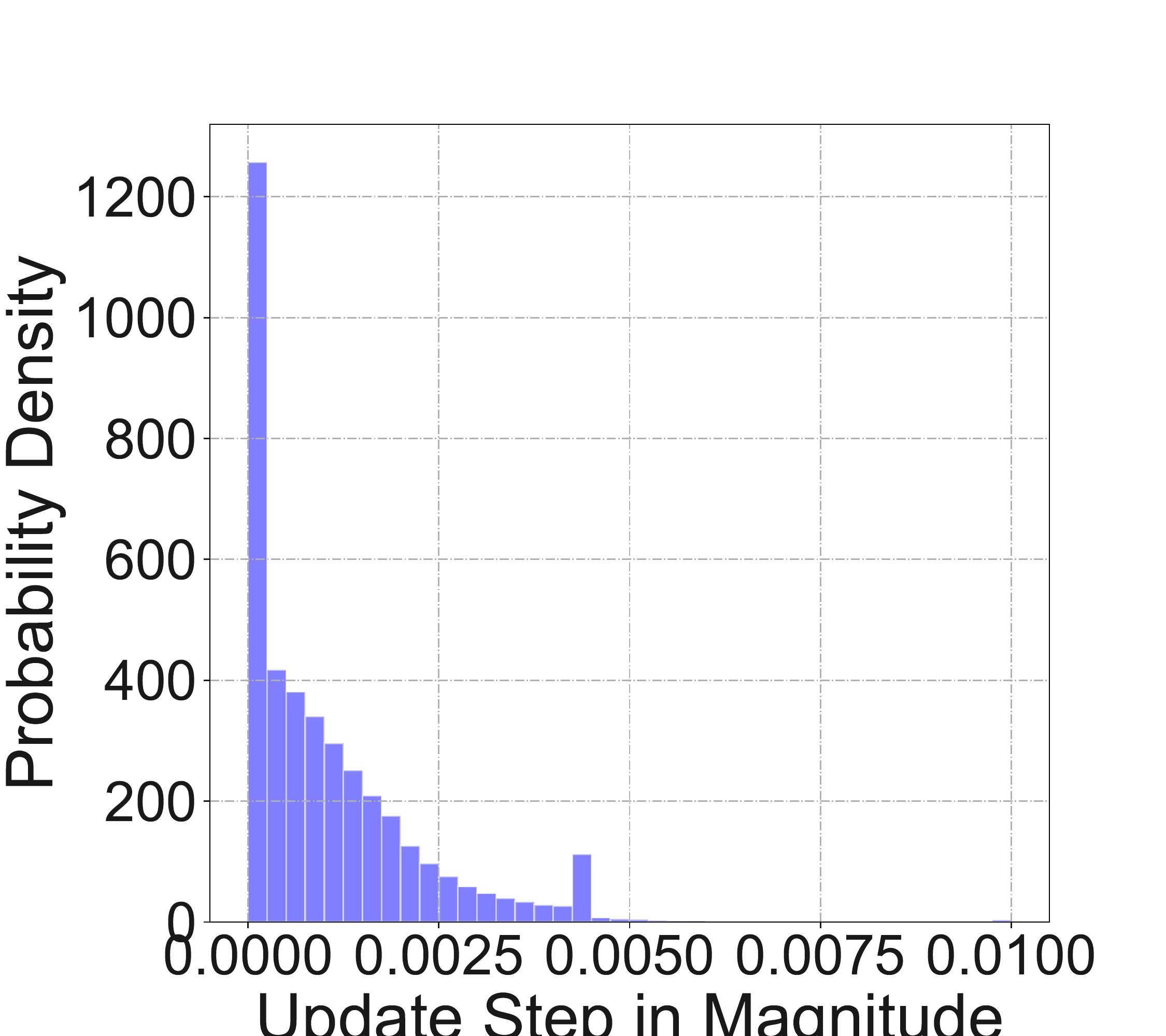} \label{fig:mnist_er_dist_step1}}
	\subfigure[Epoch 5]{\includegraphics[width=0.21\textwidth]{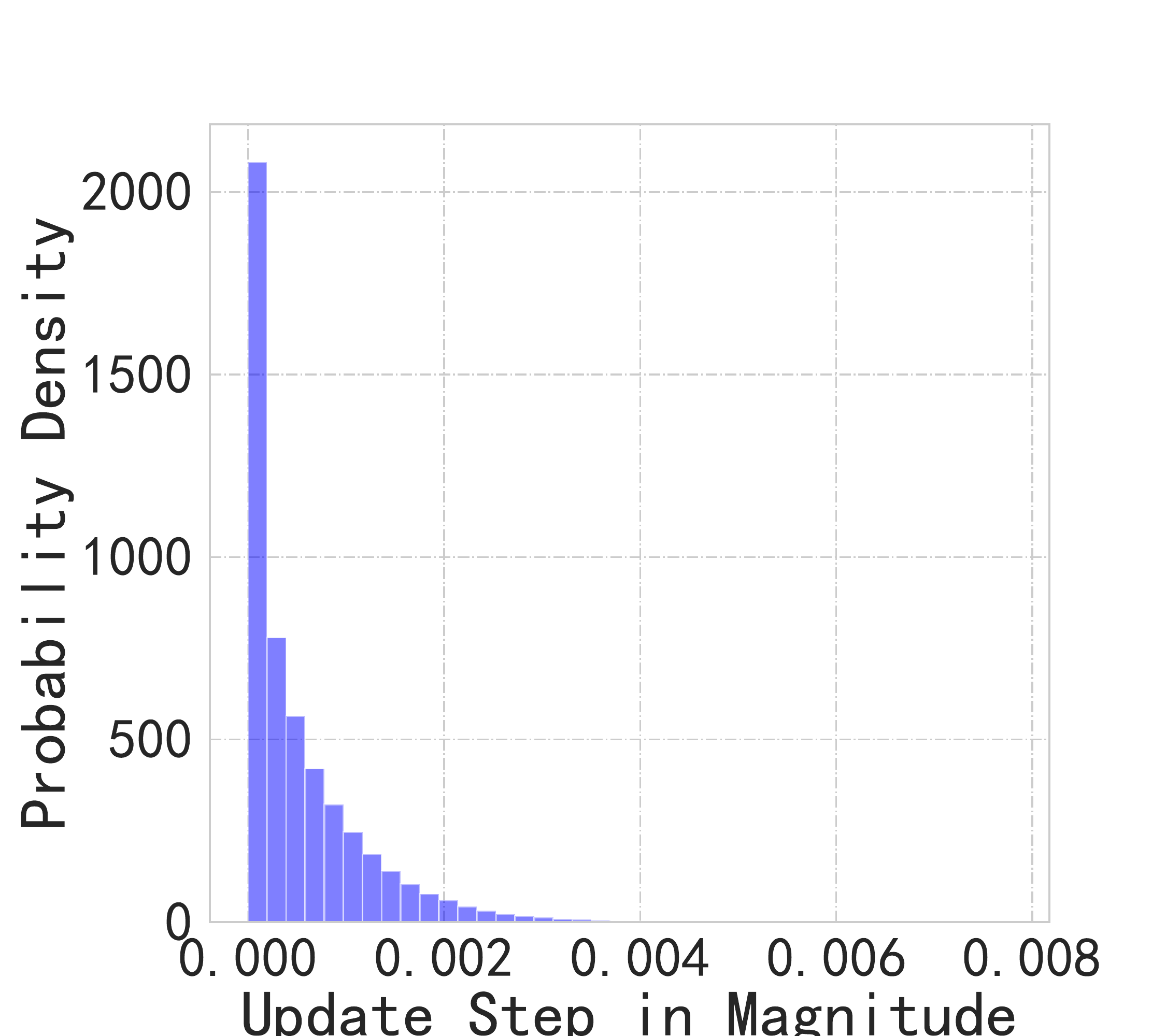} \label{fig:mnist_er_dist_step2}}
	\subfigure[Epoch 20]{\includegraphics[width=0.21\textwidth]{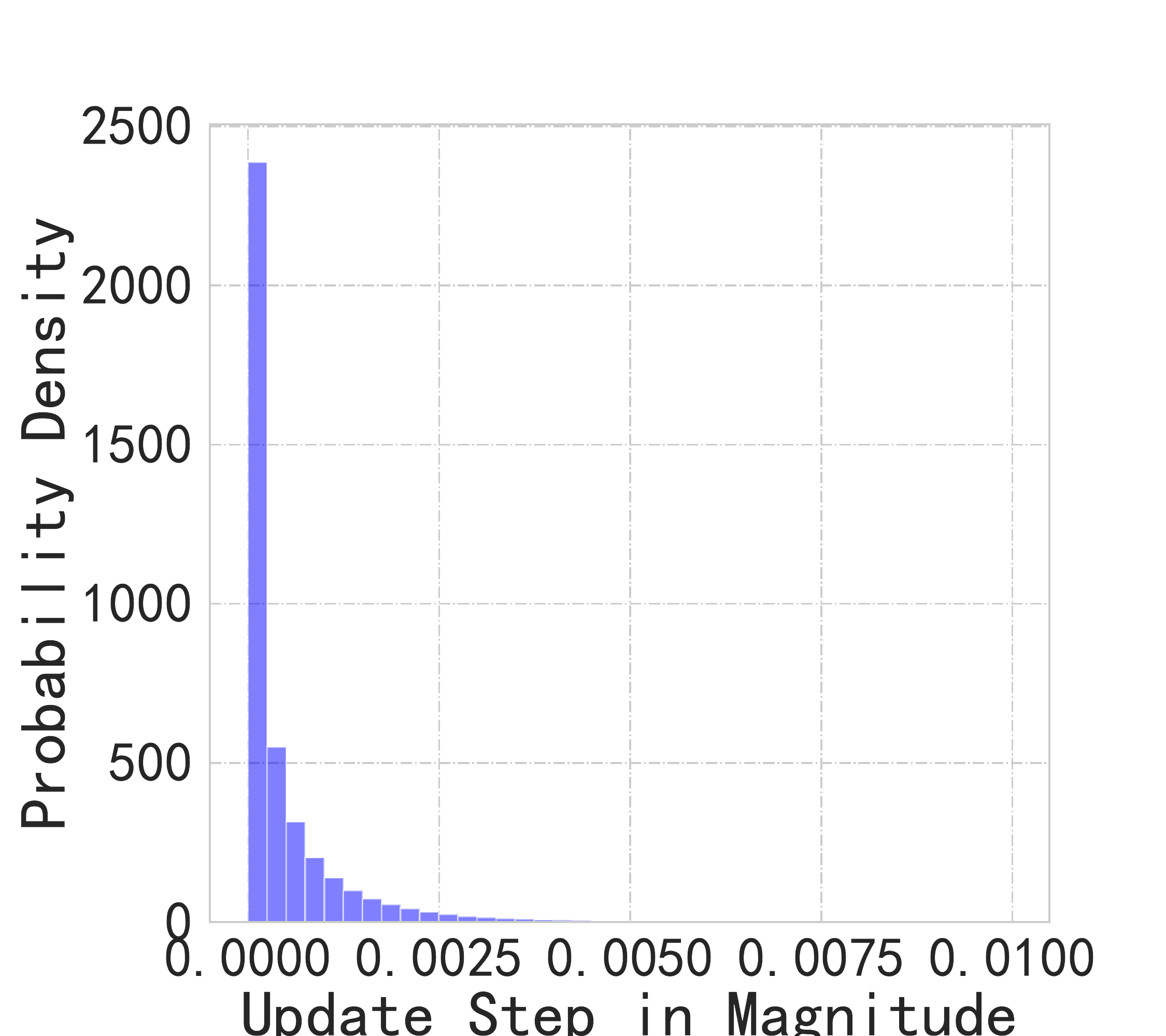} \label{fig:mnist_er_dist_step3}}
	\subfigure[Epoch 50]{\includegraphics[width=0.21\textwidth]{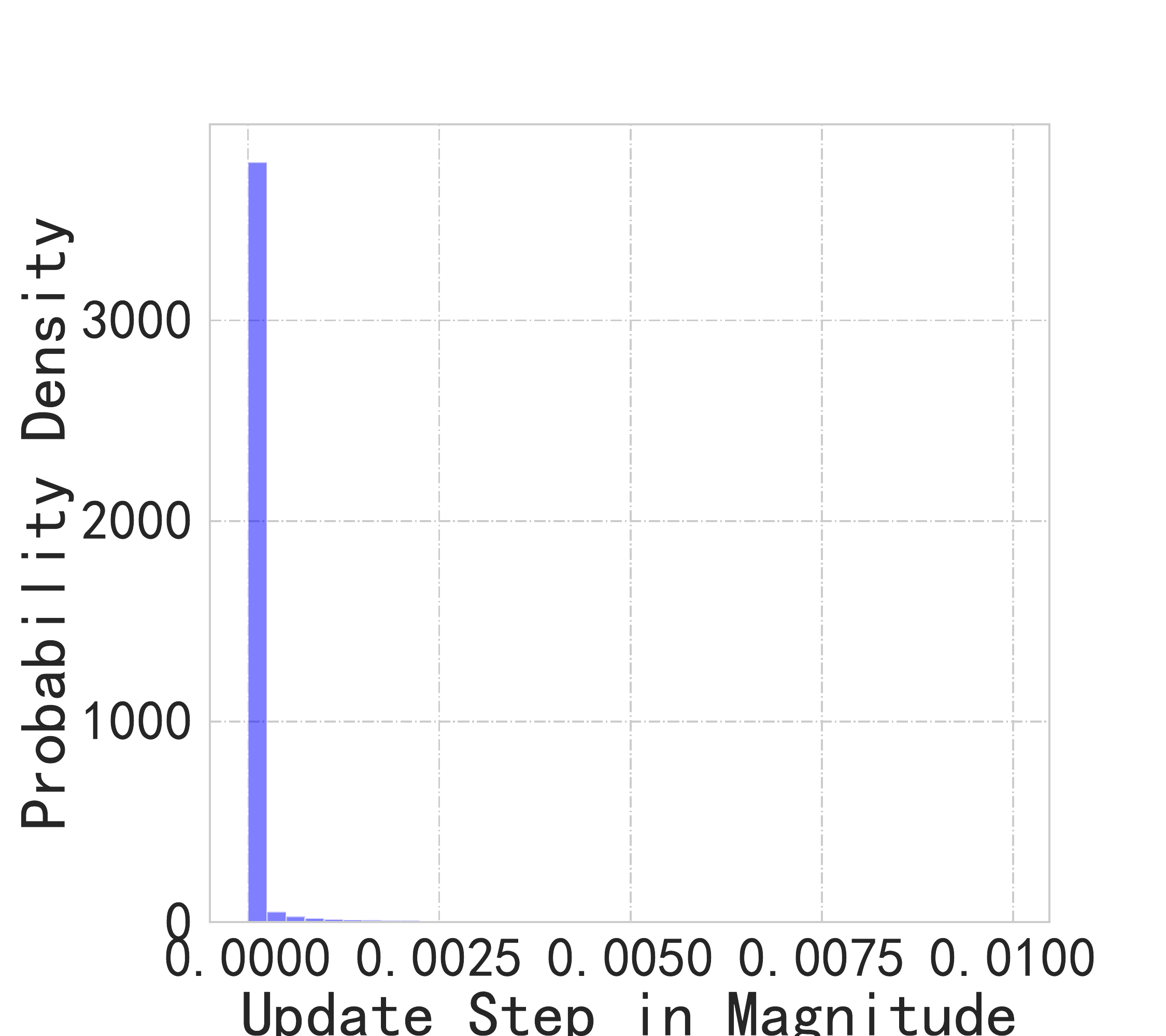} \label{fig:mnist_er_dist_step4}}
	\caption{\textbf{CNN:} The distribution of step $\Delta \bx_t$ in magnitude for different epochs in AdaSmooth ($\rho_1=0.5, \rho_2=0.99$) model.}
	\label{fig:mnist_er_dist_step}
\end{figure}

\subsection{Experiment: Convolutional Neural Networks}
Convolutional neural networks (CNN) are powerful models with non-convex objective functions. CNN with several layers of convolution, pooling and nonlinear units have mostly demonstrated remarkable success in computer vision tasks, e.g., face identification, traffic sign detection, and medical picture segmentation \citep{krizhevsky2012imagenet} and speech recognition \citep{hinton2012deep, graves2013speech}; which is partly from the local connectivity of the convolutional layers, and the rotational and shift invariance due to the pooling layers.
To evaluate the strategy and demonstrate the main advantages of the proposed AdaSmooth method, a real handwritten digit classification task, MNIST is used. For comparison with \citet{zeiler2012adadelta}'s method, we train with Relu nonlinearities and 2 convolutional layers in the front of the structure, followed by two fully connected layers. Again, the dropout with 50\% noise is adopted in the network to prevent from overfitting.

To be more concrete, the detailed architecture for each convolutional layer is described
by C$(\langle\textit{kernel size}\rangle:\langle\textit{num outputs}\rangle:\langle\textit{activation function}\rangle)$; for each fully connected
layer is described by F$(\langle \textit{num outputs} \rangle:\langle \textit{activation function} \rangle)$; for a max pooling layer is
described by MP$(\langle \textit{kernel size} \rangle:\langle \textit{stride number} \rangle)$; and for a dropout layer is described by
DP$(\langle \textit{rate} \rangle)$. Then the network structure we use can be described as follows:
\begin{equation}
\begin{aligned}
&\text{C(5:10:Relu)MP(2:2) C(5:20:Relu)MP(2:2)} \\
&\text{-
	DP(0.5) F(50:Relu)DP(0.5)F(10:Softmax)}.
\end{aligned}
\end{equation}
All methods are trained on mini-batches of 64 images per batch for 50 epochs through the training set. Setting the hyper-parameters to $\epsilon=1e-6$ and $\rho_1=0.5, \rho_2=0.9 \text{ or }0.99$, i.e., the number of periods chosen find the exponential moving average are between $3$ and 19, or between 3 and 199 iterations as discussed in Section~\ref{section:adawin} and \ref{section:adaer}. The AdaSmooth with $\rho_2=0.9$ or $0.99$ is a wide range of upper bound on the decay constant from this context; and we shall see the AdaSmooth results are not sensitive to these choices.
If not specially described, a small
learning rate $\eta=0.001$ for the convolutional networks is used in our experiments when applying stochastic descent. While again, the learning rate for the AdaDelta method is set to 1 as suggested by \citet{zeiler2012adadelta} and for the AdaSmoothDelta method is set to 0.5 as discussed in Section~\ref{section:adasmoothdelta}.

In Figure~\ref{fig:mnist_loss_train}, we compare SGD with Momentum, AdaGrad, RMSProp, AdaDelta, AdaSmooth, and AdaSmoothDelta in optimizing the training set loss (negative log likelihood). 
%The unaltered SGD method does the worst in this case, whereas adding momentum term to it significantly improves performance. 
The RMSProp ($\rho=0.9$) does the worst in this case, whereas tuning the decay constant to $\rho=0.99$ can significantly improve performance making the RMSProp sensitive to the hyper-parameter (hence the learning rates per-dimension).
Further, though the training loss of AdaDelta ($\rho=0.9$) decreases fastest in the first 3 epochs, the performance becomes poor at the end of the training, with an average loss larger than 0.15; while the overall performance of AdaDelta ($\rho=0.99$) works better than the former, however, its overall accuracy is still worse than AdaSmooth. This also reveals the same drawback of AdaGrad for the AdaDelta method, i.e., they are sensitive to the choices of hyper-parameters.

In order to evaluate whether the AdaSmooth actually can find the compensation we want, we also explore a random selection for the number of periods/iterations $N$, termed as \textit{AdaSmooth(Random)} in the sequel, which selects the decay constant between $\rho_1$ and $\rho_2$ randomly during different batches. The result shows the AdaSmooth(Random) cannot find the adaptive learning rates per-dimension in the right way showing our method actually finds the correct choices between the lower and upper bound of the decay constant.

The analysis of different methods in test loss and test accuracy are consistent with that of training loss as shown in Figure~\ref{fig:mnist_loss_test} and Figure~\ref{fig:mnist_acc_test}. We further save the effective ratios in magnitude as stated in Eq~\eqref{eqution:signoiase-er-delta} during each epoch; the distributions of them are shown in Figure~\ref{fig:mnist_er_dist}. In the first few epochs, the ERs in magnitude have more weights in large values, while in the last few epochs, the probability decreases into a stationary distribution; in other words, the ER distributions in the last few epochs are similar. In Figure~\ref{fig:mnist_er_dist_thres}, we put a threshold on the ER values, the percentages of ERs larger than the threshold are plotted; when the training is in progress, the ERs tend to approach smaller values indicating the algorithm goes into stationary points, the (local) minima. While Figure~\ref{fig:mnist_er_dist_step} shows the absolute values of update step $\Delta \bx_t$ for different epochs; when the training is in the later stage, the parameters are closer to the (local) minima, indicating a smaller update step and favoring a relatively smaller period (resulting from both the step and the direction of the movement) of exponential moving average in Eq~\eqref{equation:squared-ssc}, which in turn compensates relatively less for the learning rate per-dimension and makes the update faster. 
%When closing to these stationary points, AdaSmooth tends to use a larger periods (i.e., a larger decay constant) to compensate the learning rate per-dimension.
In Table~\ref{fig:cnnn_adasmoother_testeval}, we further show test accuracy for AdaSmooth and AdaSmoothDelta with various hyper-parameters after 10 epochs in which case different choices of the parameters do not significantly alter performance.

\begin{table}[!h]
\begin{tabular}{ll}
\hline
Method & MNIST \\	 \hline
AdaGrad ($\eta$=0.01) & 96.82\%\\
AdaGrad ($\eta$=0.001) & 89.11\%\\
RMSProp ($\rho=0.99$) & 97.82\%\\
RMSProp ($\rho=0.9$) & 97.88\%\\
AdaDelta ($\rho=0.99$) & 97.83\%\\
AdaDelta ($\rho=0.9$) & 98.20\%\\
AdaSmooth ($\rho_1=0.5, \rho_2=0.9$) & 98.13\%\\
AdaSmooth ($\rho_1=0.5, \rho_2=0.95$) & 98.12\%\\
AdaSmooth ($\rho_1=0.5, \rho_2=0.99$) & 98.12\%\\
AdaSmoDel. ($\rho_1=0.5, \rho_2=0.9, \eta=0.5$) & 98.86\%\\
AdaSmoDel. ($\rho_1=0.5, \rho_2=0.95, \eta=0.5$) & 98.91\%\\
AdaSmoDel. ($\rho_1=0.5, \rho_2=0.99, \eta=0.5$) & 98.78\%\\
AdaSmoDel. ($\rho_1=0.5, \rho_2=0.99, \eta=0.6$) & 98.66\%\\
AdaSmoDel. ($\rho_1=0.5, \rho_2=0.99, \eta=0.7$) & 98.66\%\\
AdaSmoDel. ($\rho_1=0.5, \rho_2=0.99, \eta=0.8$) & 98.58\%\\
\hline
\end{tabular}
\caption{\textbf{CNN}: Best out-of-sample evaluation in test accuracy for AdaSmooth and AdaSmoothDelta with various hyper-parameters after 10 epochs. AdaSmDel is short for AdaSmoothDelta.}
\label{fig:cnnn_adasmoother_testeval}
\end{table}

\begin{figure}[!h]
	\centering
	\subfigure[MNIST training loss]{\includegraphics[width=0.491\textwidth, ]{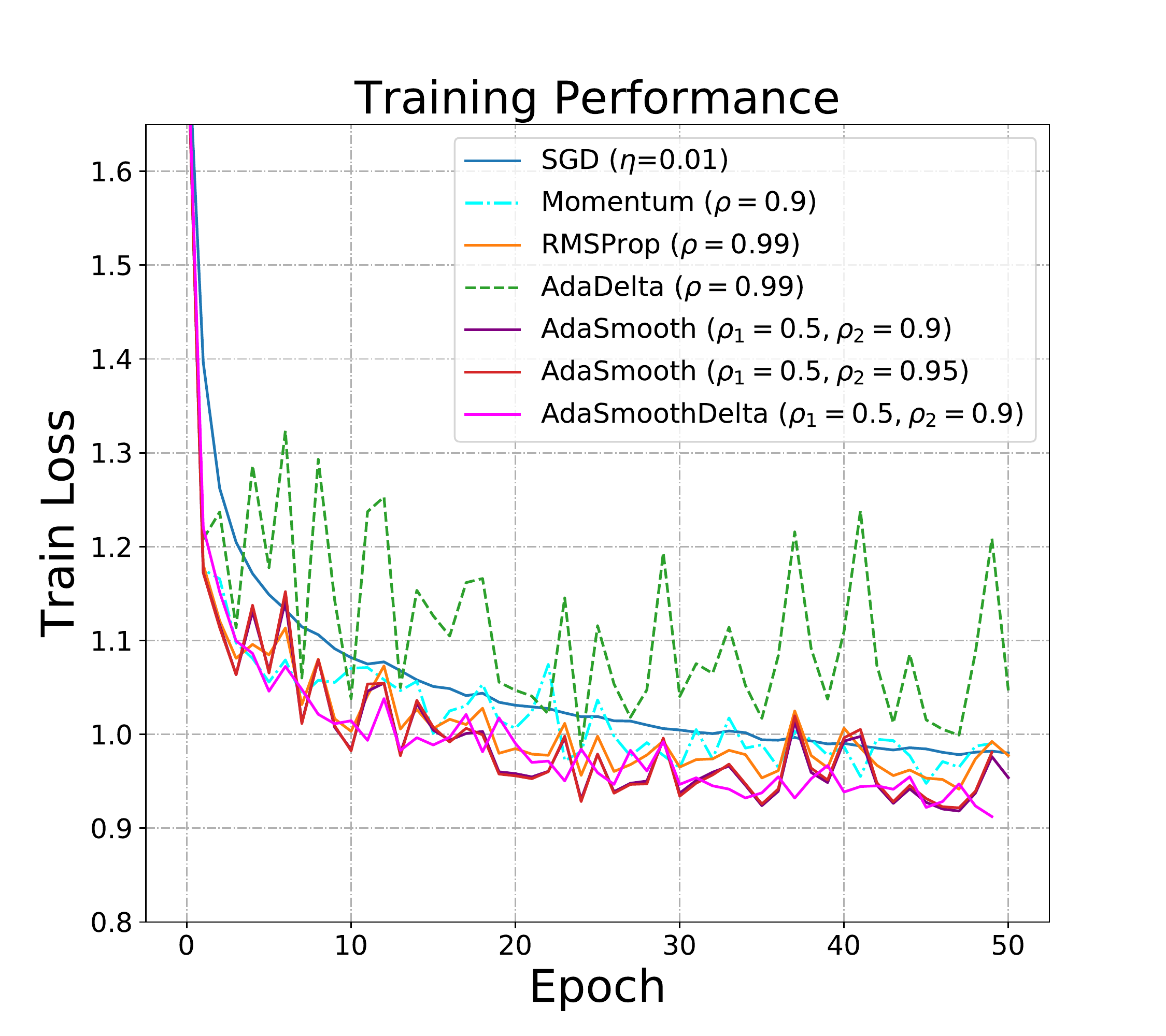} \label{fig:mnist_loss_logis_train}}
	\subfigure[Census Income training loss]{\includegraphics[width=0.491\textwidth]{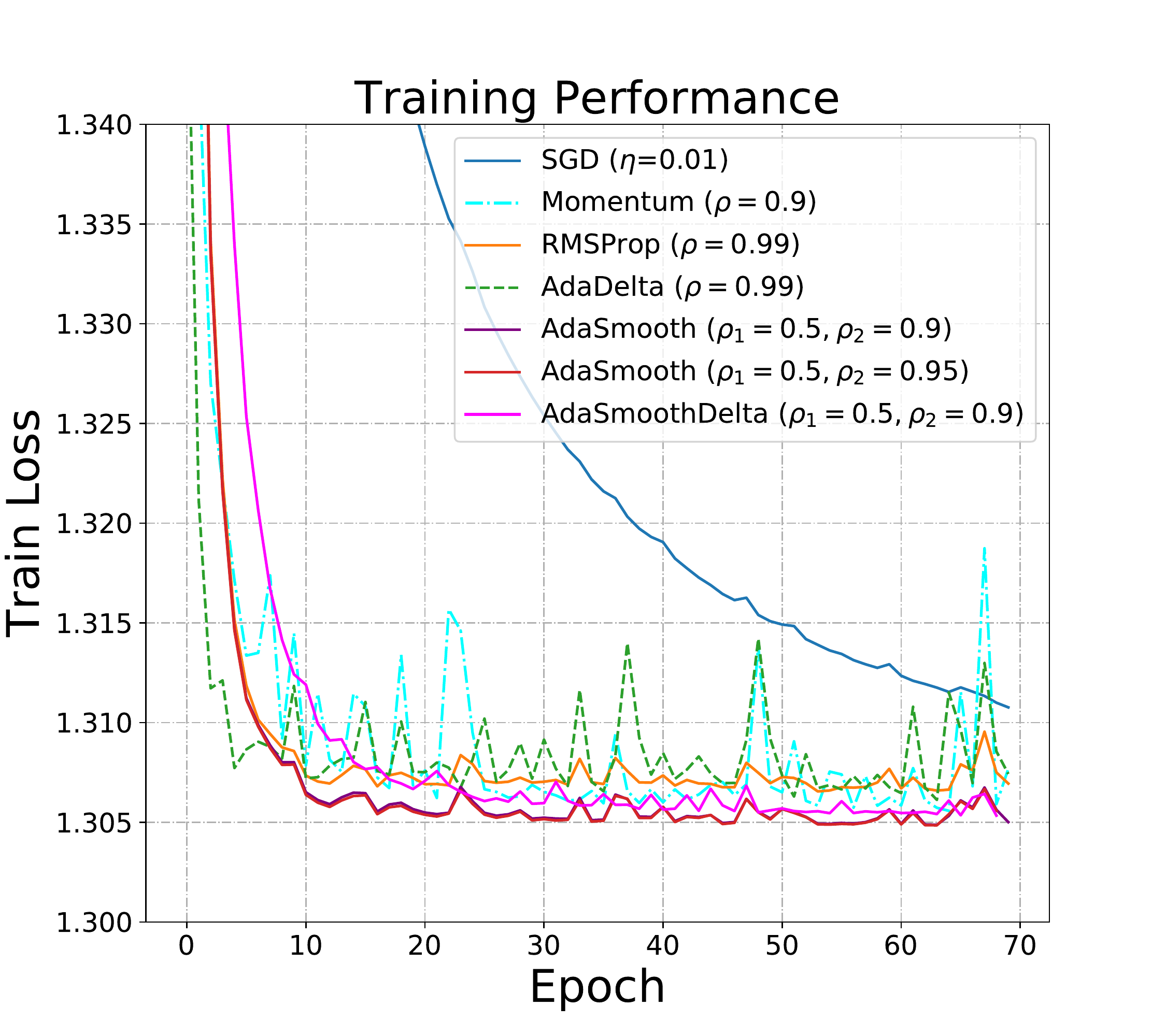} \label{fig:census_loss_logis_train}}
	\caption{\textbf{Logistic regression:} Comparison of descent methods on MNIST digit and Census Income data sets for 50 and 70 epochs with logistic regression. Notice that the training curve of AdaSmooth $(\rho_1=0.5,\rho_2=0.9)$ is close to $(\rho_1=0.5,\rho_2=0.95)$ in this case for both data sets.}
	\label{fig:mnist_census_loss_logistic}
\end{figure}

\subsection{Experiment: Logistic Regression}
Logistic regression is an analytic technique for multivariate modeling of categorical dependent variables
and has a well-studied convex objective, making it suitable for comparison of different optimizers without worrying about going into local minima \citep{menard2002applied}. 
To evaluate, again if not mentioned explicitly, the global learning rates are set to $\eta=0.001$ in all scenarios. While a relatively large learning rate ($\eta=0.01$) is used for AdaGrad method since its accumulated decaying effect; learning rate for the AdaDelta method is set to 1 as suggested by \citet{zeiler2012adadelta} and for the AdaSmoothDelta method is set to 0.5 as discussed in Section~\ref{section:adasmoothdelta}.
The loss curves for training processes are shown in Figure~\ref{fig:mnist_loss_logis_train} and ~\ref{fig:census_loss_logis_train}, where we compare SGD, SGD with Momentum, RMSProp, AdaDelta, AdaSmooth, and AdaSmoothDelta in optimizing the training set losses for MNIST and Census Income data sets respectively. The unaltered SGD method does the worst in this case. AdaSmooth performs slightly better than RMSProp in the MNIST case and much better than the latter in the Census Income case. 
AdaSmooth matches the fast convergence of AdaDelta (in which case AdaSmooth converges slightly slower than AdaDelta in the first few epochs), while
AdaSmooth continues to reduce the training loss, converging to the best performance in these models.
As aforementioned, when $\rho_1=\rho_2=0.9$, the AdaSmooth recovers to RMSProp with $\rho=0.99$ (so as the AdaSmoothDelta and AdaDelta case). In all cases, the AdaSmooth results perform better while the difference between various hyper-parameters for AdaSmooth is not significant; there are almost no differences in AdaSmooth results with different hyper-parameter settings, indicating its insensitivity to hyper-parameters.
In Table~\ref{fig:logistic_table_perform_logistic}, we compare the training set accuracy of various algorithms; AdaSmooth works best in the MNIST case, while the superiority in Census Income data is not significant. Similar results as the MLP scenario can be observed in the test accuracy and we shall not give the details for simplicity.

\begin{table}[!h]
\begin{tabular}{lll}
\hline
Method & MNIST &  Census \\ \hline
SGD ($\eta$=0.01) & 93.29\% & 84.84\%\\
Momentum ($\rho=0.9$) & 93.39\%& 84.94\%\\
RMSProp ($\rho=0.99$) & 93.70\% & 84.94\%\\
AdaDelta ($\rho=0.99$) & 93.48\%& 84.94\%\\
AdaSmooth ($\rho_1=0.5, \rho_2=0.9$) & \textbf{93.74}\%& 84.92\%\\
AdaSmooth ($\rho_1=0.5, \rho_2=0.95$) & 93.71\%& 84.94\%\\
AdaSmoDel. ($\rho_1=0.5, \rho_2=0.9$) & 93.66\%& \textbf{84.97}\%\\
\hline
\end{tabular}
\caption{\textbf{Logistic regression}: Best in-sample evaluation in training accuracy. AdaSmoDel is short for AdaSmoothDelta.}
\label{fig:logistic_table_perform_logistic}
\end{table}

\section{Conclusion}
The aim of this paper is to solve the hyper-parameter tuning in the gradient-based optimization methods for machine learning problems with large data sets and high-dimensional parameter spaces.
We propose a simple and computationally efficient algorithm that requires little memory and is easy to implement for gradient-based optimization of stochastic objective functions. 
%To evaluate the proposed algorithm, real data sets including MNIST and Census Income are used in empirical analysis. 
Overall, we show that AdaSmooth is a versatile algorithm that scales to large-scale high-dimensional machine learning problems and AdaSmoothDelta is an algorithm insensitive to both the global learning rate and the hyper-parameters with caution to set the global learning rate smaller than 1.

\small
\bibliography{bib}

\begin{thebibliography}{30}
\providecommand{\natexlab}[1]{#1}
\providecommand{\url}[1]{\texttt{#1}}
\expandafter\ifx\csname urlstyle\endcsname\relax
  \providecommand{\doi}[1]{doi: #1}\else
  \providecommand{\doi}{doi: \begingroup \urlstyle{rm}\Url}\fi

\bibitem[Becker \& Le~Cun(1988)Becker and Le~Cun]{becker1988improving}
Becker, Sue and Le~Cun, Yann.
\newblock Improving the convergence of back-propagation learning with.
\newblock 1988.

\bibitem[Dauphin et~al.(2014)Dauphin, Pascanu, Gulcehre, Cho, Ganguli, and
  Bengio]{dauphin2014identifying}
Dauphin, Yann~N, Pascanu, Razvan, Gulcehre, Caglar, Cho, Kyunghyun, Ganguli,
  Surya, and Bengio, Yoshua.
\newblock Identifying and attacking the saddle point problem in
  high-dimensional non-convex optimization.
\newblock \emph{Advances in neural information processing systems}, 27, 2014.

\bibitem[Dozat(2016)]{dozat2016incorporating}
Dozat, Timothy.
\newblock Incorporating nesterov momentum into adam.
\newblock 2016.

\bibitem[Duchi et~al.(2011)Duchi, Hazan, and Singer]{duchi2011adaptive}
Duchi, John, Hazan, Elad, and Singer, Yoram.
\newblock Adaptive subgradient methods for online learning and stochastic
  optimization.
\newblock \emph{Journal of machine learning research}, 12\penalty0 (7), 2011.

\bibitem[Graves et~al.(2013)Graves, Mohamed, and Hinton]{graves2013speech}
Graves, Alex, Mohamed, Abdel-rahman, and Hinton, Geoffrey.
\newblock Speech recognition with deep recurrent neural networks.
\newblock In \emph{2013 IEEE international conference on acoustics, speech and
  signal processing}, pp.\  6645--6649. Ieee, 2013.

\bibitem[Hinton et~al.(2012{\natexlab{a}})Hinton, Deng, Yu, Dahl, Mohamed,
  Jaitly, Senior, Vanhoucke, Nguyen, Sainath, et~al.]{hinton2012deep}
Hinton, Geoffrey, Deng, Li, Yu, Dong, Dahl, George~E, Mohamed, Abdel-rahman,
  Jaitly, Navdeep, Senior, Andrew, Vanhoucke, Vincent, Nguyen, Patrick,
  Sainath, Tara~N, et~al.
\newblock Deep neural networks for acoustic modeling in speech recognition: The
  shared views of four research groups.
\newblock \emph{IEEE Signal processing magazine}, 29\penalty0 (6):\penalty0
  82--97, 2012{\natexlab{a}}.

\bibitem[Hinton et~al.(2012{\natexlab{b}})Hinton, Srivastava, and
  Swersky]{hinton2012neural}
Hinton, Geoffrey, Srivastava, Nitish, and Swersky, Kevin.
\newblock Neural networks for machine learning lecture 6a overview of
  mini-batch gradient descent.
\newblock \emph{Cited on}, 14\penalty0 (8):\penalty0 2, 2012{\natexlab{b}}.

\bibitem[Kaufman(1995)]{kaufman1995smarter}
Kaufman, Perry~J.
\newblock Smarter trading, 1995.

\bibitem[Kaufman(2013)]{kaufman2013trading}
Kaufman, Perry~J.
\newblock \emph{Trading Systems and Methods,+ Website}, volume 591.
\newblock John Wiley \& Sons, 2013.

\bibitem[Kingma \& Ba(2014)Kingma and Ba]{kingma2014adam}
Kingma, Diederik~P and Ba, Jimmy.
\newblock Adam: A method for stochastic optimization.
\newblock \emph{arXiv preprint arXiv:1412.6980}, 2014.

\bibitem[Krizhevsky et~al.(2012)Krizhevsky, Sutskever, and
  Hinton]{krizhevsky2012imagenet}
Krizhevsky, Alex, Sutskever, Ilya, and Hinton, Geoffrey~E.
\newblock Imagenet classification with deep convolutional neural networks.
\newblock \emph{Advances in neural information processing systems}, 25, 2012.

\bibitem[Le~Roux \& Fitzgibbon(2010)Le~Roux and Fitzgibbon]{le2010fast}
Le~Roux, Nicolas and Fitzgibbon, Andrew~W.
\newblock A fast natural newton method.
\newblock In \emph{ICML}, 2010.

\bibitem[LeCun(1998)]{lecun1998mnist}
LeCun, Yann.
\newblock The {MNIST} database of handwritten digits.
\newblock \emph{http://yann. lecun. com/exdb/mnist/}, 1998.

\bibitem[Lu(2022{\natexlab{a}})]{lu2022exploring}
Lu, Jun.
\newblock Exploring classic quantitative strategies.
\newblock \emph{arXiv preprint arXiv:2202.11309}, 2022{\natexlab{a}}.

\bibitem[Lu(2022{\natexlab{b}})]{lu2022matrix}
Lu, Jun.
\newblock Matrix decomposition and applications.
\newblock \emph{arXiv preprint arXiv:2201.00145}, 2022{\natexlab{b}}.

\bibitem[Lu \& Yi(2022)Lu and Yi]{lu2022reducing}
Lu, Jun and Yi, Shao.
\newblock Reducing overestimating and underestimating volatility via the
  augmented blending-{ARCH} model.
\newblock \emph{arXiv preprint arXiv:2203.12456}, 2022.

\bibitem[Menard(2002)]{menard2002applied}
Menard, Scott.
\newblock \emph{Applied logistic regression analysis}, volume 106.
\newblock Sage, 2002.

\bibitem[Moulines \& Bach(2011)Moulines and Bach]{moulines2011non}
Moulines, Eric and Bach, Francis.
\newblock Non-asymptotic analysis of stochastic approximation algorithms for
  machine learning.
\newblock \emph{Advances in neural information processing systems}, 24, 2011.

\bibitem[Polyak \& Juditsky(1992)Polyak and Juditsky]{polyak1992acceleration}
Polyak, Boris~T and Juditsky, Anatoli~B.
\newblock Acceleration of stochastic approximation by averaging.
\newblock \emph{SIAM journal on control and optimization}, 30\penalty0
  (4):\penalty0 838--855, 1992.

\bibitem[Qian(1999)]{qian1999momentum}
Qian, Ning.
\newblock On the momentum term in gradient descent learning algorithms.
\newblock \emph{Neural networks}, 12\penalty0 (1):\penalty0 145--151, 1999.

\bibitem[Robbins \& Monro(1951)Robbins and Monro]{robbins1951stochastic}
Robbins, Herbert and Monro, Sutton.
\newblock A stochastic approximation method.
\newblock \emph{The annals of mathematical statistics}, pp.\  400--407, 1951.

\bibitem[Ruder(2016)]{ruder2016overview}
Ruder, Sebastian.
\newblock An overview of gradient descent optimization algorithms.
\newblock \emph{arXiv preprint arXiv:1609.04747}, 2016.

\bibitem[Rumelhart et~al.(1986)Rumelhart, Hinton, and
  Williams]{rumelhart1986learning}
Rumelhart, David~E, Hinton, Geoffrey~E, and Williams, Ronald~J.
\newblock Learning representations by back-propagating errors.
\newblock \emph{nature}, 323\penalty0 (6088):\penalty0 533--536, 1986.

\bibitem[Ruppert(1988)]{ruppert1988efficient}
Ruppert, David.
\newblock Efficient estimations from a slowly convergent robbins-monro process.
\newblock Technical report, Cornell University Operations Research and
  Industrial Engineering, 1988.

\bibitem[Schaul et~al.(2013)Schaul, Zhang, and LeCun]{schaul2013no}
Schaul, Tom, Zhang, Sixin, and LeCun, Yann.
\newblock No more pesky learning rates.
\newblock In \emph{International conference on machine learning}, pp.\
  343--351. PMLR, 2013.

\bibitem[Smith(2017)]{smith2017cyclical}
Smith, Leslie~N.
\newblock Cyclical learning rates for training neural networks.
\newblock In \emph{2017 IEEE winter conference on applications of computer
  vision (WACV)}, pp.\  464--472. IEEE, 2017.

\bibitem[Srivastava et~al.(2014)Srivastava, Hinton, Krizhevsky, Sutskever, and
  Salakhutdinov]{srivastava2014dropout}
Srivastava, Nitish, Hinton, Geoffrey, Krizhevsky, Alex, Sutskever, Ilya, and
  Salakhutdinov, Ruslan.
\newblock Dropout: a simple way to prevent neural networks from overfitting.
\newblock \emph{The journal of machine learning research}, 15\penalty0
  (1):\penalty0 1929--1958, 2014.

\bibitem[Sutskever et~al.(2013)Sutskever, Martens, Dahl, and
  Hinton]{sutskever2013importance}
Sutskever, Ilya, Martens, James, Dahl, George, and Hinton, Geoffrey.
\newblock On the importance of initialization and momentum in deep learning.
\newblock In \emph{International conference on machine learning}, pp.\
  1139--1147. PMLR, 2013.

\bibitem[You et~al.(2019)You, Li, Reddi, Hseu, Kumar, Bhojanapalli, Song,
  Demmel, Keutzer, and Hsieh]{you2019large}
You, Yang, Li, Jing, Reddi, Sashank, Hseu, Jonathan, Kumar, Sanjiv,
  Bhojanapalli, Srinadh, Song, Xiaodan, Demmel, James, Keutzer, Kurt, and
  Hsieh, Cho-Jui.
\newblock Large batch optimization for deep learning: Training bert in 76
  minutes.
\newblock \emph{arXiv preprint arXiv:1904.00962}, 2019.

\bibitem[Zeiler(2012)]{zeiler2012adadelta}
Zeiler, Matthew~D.
\newblock Adadelta: an adaptive learning rate method.
\newblock \emph{arXiv preprint arXiv:1212.5701}, 2012.

\end{thebibliography}
\bibliographystyle{sty}

%\pagebreak
\appendix
%\section{Extentio}
%\begin{python}
%# get fast SC and slow SC
%fastsc = 1 - group['rho1']
%slowsc = 1 - group['rho2']
%# cal ER
%er = torch.abs(state['signal']) / (state['noise'] + group['eps'])
%ssc = er * (fastsc - slowsc) + slowsc
%ssc = ssc * ssc
%state['sum'] = (1 - ssc) * state['sum'] + ssc * grad ** 2
%std = state['sum'].sqrt().add_(group['eps'])
%
%# get update step and update
%delta = -clr*grad/std
%p.data = p.data +delta
%
%# update noise and signal vector
%state['signal'] = state['signal'] + delta
%state['noise'] = state['noise'] + torch.abs(delta)
%\end{python}

\end{document}